\documentclass[10pt,twocolumn,letterpaper]{article}

\usepackage[pagenumbers]{cvpr} 
\usepackage{graphicx}
\usepackage{amsmath}
\usepackage{amssymb}
\usepackage{booktabs}
\usepackage{flushend}
\usepackage[pagebackref,breaklinks,colorlinks]{hyperref}

\usepackage[capitalize]{cleveref}
\crefname{section}{Sec.}{Secs.}
\Crefname{section}{Section}{Sections}
\Crefname{table}{Table}{Tables}
\crefname{table}{Tab.}{Tabs.}

\usepackage{pifont}
\usepackage{multirow}  
\graphicspath{{figures/}}  
\newcommand{\tabincell}[2]{\begin{tabular}{@{}#1@{}}#2\end{tabular}} 

\begin{document}
	
	\title{Noise Self-Regression: A New Learning Paradigm to Enhance Low-Light Images Without Task-Related Data}
	
	\author{Zhao~Zhang$^{1}$, Suiyi~Zhao$^{1*}$, Xiaojie~Jin$^2$, Mingliang~Xu$^3$, Yi Yang$^4$, Shuicheng Yan$^5$ and Meng Wang$^1*$ \vspace{2mm}\\
		$^{1}$ Hefei University of Technology, Hefei, China\\
		$^{2}$ Bytedance Research, USA\\
		$^{3}$ Zhengzhou University, Zhengzhou, China\\
		$^{4}$ Zhejiang University, Hangzhou, China\\
		$^{5}$ Kunlun 2050 Research \& Skywork AI, Singapore
	}

\maketitle

\begin{abstract}
	\vspace{-4mm}
	
	Deep learning-based low-light image enhancement (LLIE) is a task of leveraging deep neural networks to enhance the image illumination while keeping the image content unchanged. From the perspective of training data, existing methods complete the LLIE task driven by one of the following three data types: paired data, unpaired data and zero-reference data. Each type of these data-driven methods has its own advantages, e.g., zero-reference data-based methods have very low requirements on training data and can meet the human needs in many scenarios. In this paper, we leverage pure Gaussian noise to complete the LLIE task, which further reduces the requirements for training data in LLIE tasks and can be used as another alternative in practical use. Specifically, we propose Noise SElf-Regression (NoiSER) without access to any task-related data, simply learns a convolutional neural network equipped with an instance-normalization layer by taking a random noise image, $\mathcal{N}(0,\sigma^2)$ for each pixel, as both input and output for each training pair, and then the low-light image is fed to the trained network for predicting the normal-light image. Technically, an intuitive explanation for its effectiveness is as follows: 1) the self-regression reconstructs the contrast between adjacent pixels of the input image, 2) the instance-normalization layer may naturally remediate the overall magnitude/lighting of the input image, and 3) the $\mathcal{N}(0,\sigma^2)$ assumption for each pixel enforces the output image to follow the well-known gray-world hypothesis \cite{Gary-world_Hypothesis} when the image size is big enough. Compared to current state-of-the-art LLIE methods with access to different task-related data, NoiSER is highly competitive in enhancement quality, yet with a much smaller model size, and much lower training and inference cost. In addition, the experiments also demonstrate that NoiSER has great potential in overexposure suppression and joint processing with other restoration tasks. 
\end{abstract}

\vspace{-6mm}
\section{Introduction}
\label{sec:intro}

\begin{figure}[t]
	\centering
	\vspace{-5mm}
	\includegraphics[width=0.89\columnwidth]{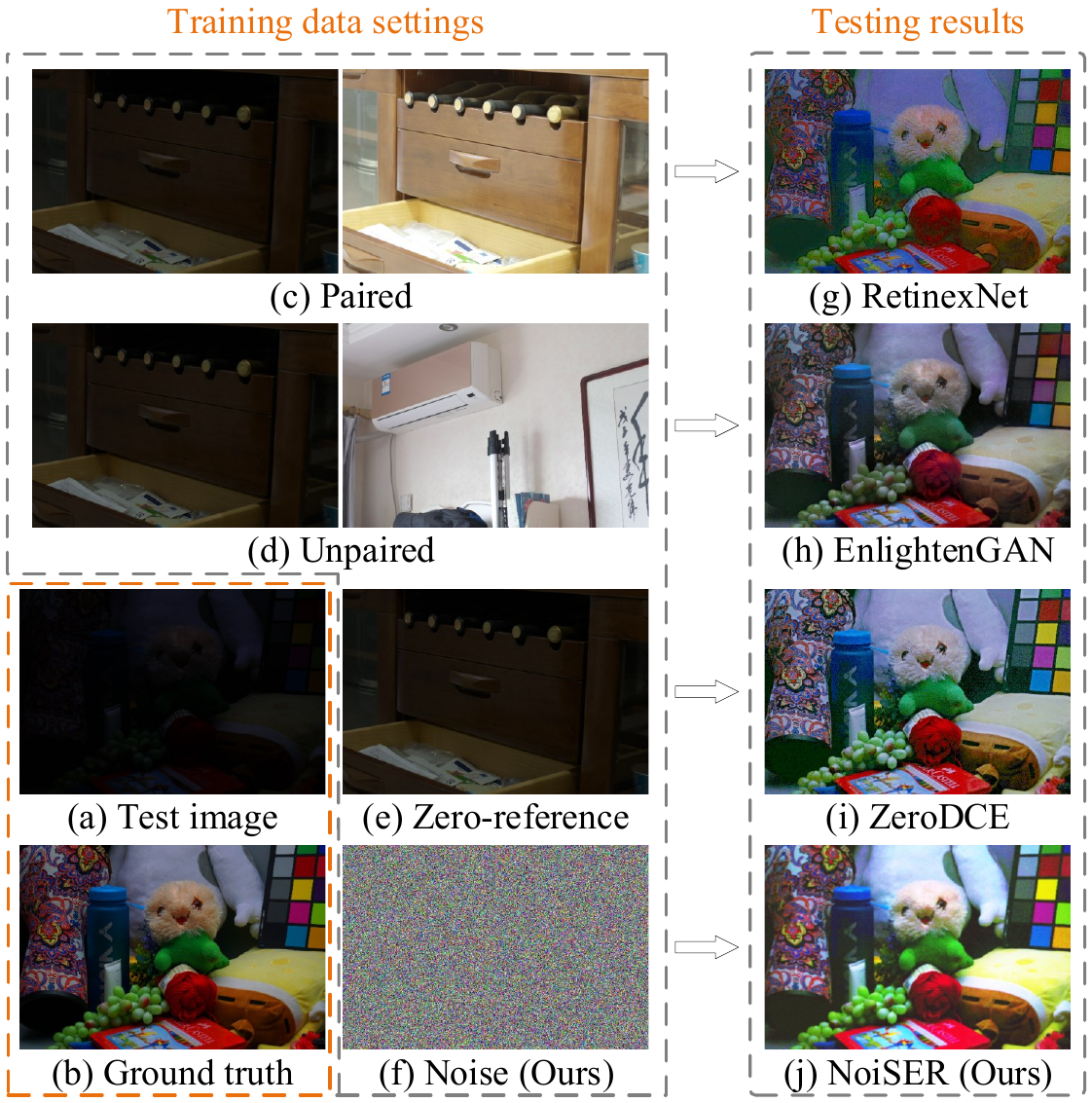}
	
	\caption{Comparison of training data settings used in current LLIE methods, including the paired data in RetinexNet \cite{RetinexNet}, unpaired data in EnlightenGAN \cite{EnlightenGAN}, zero-reference data in Zero-DCE \cite{Zero-DCE}, and the noise (task-irrelevant data) in our proposed NoiSER. Clearly, NoiSER performs the best in naturalness and detail recovery.}
	\label{fig:Motivation}
	\vspace{-3mm}
\end{figure}

Unfavorable illumination is rather common when taking photographs, and the resulted photos are often poorly illuminated, greatly hindering the understanding of their contents. Low-light image enhancement (LLIE) aims to transform low-light images into normal-light ones by refining the illumination of the images. Traditional LLIE methods are usually based on the histogram equalization (HE) \cite{HE1,HE2} or the retinex theory \cite{LIME, DUAL, LR3M, Retinex-Based1}, and they usually cause unpleasant artifacts. Recently, deep neural networks (DNNs) \cite{CRNet, DeepDeblur,DerainCycleGAN} have been successfully used for various high-level \cite{CycleGAN,Segmentation-Survey,Weakly--Supervised-Object-Detection,Novel-Captioner} and low-level \cite{FCL-GAN,MPRNet,Bayes-Unpaired-Restoration} vision tasks due to their strong learning abilities. This also gives birth to advanced deep learning-based LLIE methods \cite{Zero-DCE,Zero-DCE++,SCI,EnlightenGAN,FIDE}, in which task-related data is a necessary condition.
\begin{figure*}[t]
	\centering
	\includegraphics[width=1.85\columnwidth]{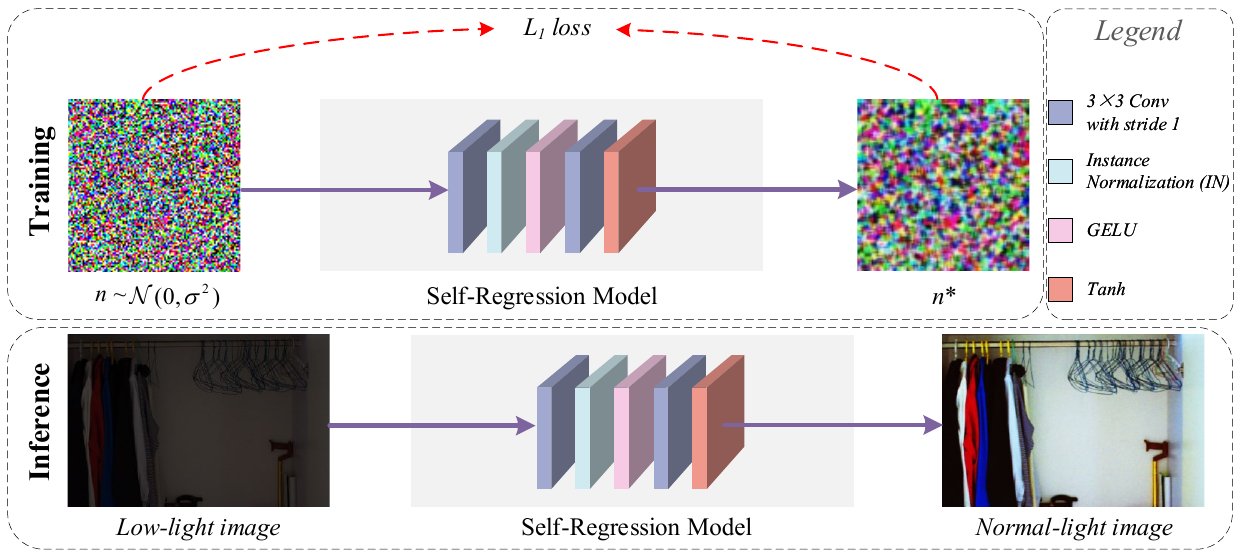}
	\vspace{-2mm}
	\caption{The training and inference pipeline of NoiSER. During training, NoiSER just samples noise $n\sim \mathcal{N}(0,\sigma^2)$ as both model input and supervised signal to train a self-regression model (SRM), i.e., $n^\ast$$=$$SRM(n)$$\approx$$n$. During inference, the trained SRM can directly enhance low-light images. Note that the designed SRM is equipped with one instance normalization and two non-linear activation layers without shortcuts, which avoids learning an identity mapping.}
	\vspace{-4mm}
	\label{fig:Architecture}
\end{figure*}

\textbf{Paired data}, including the paired low-light and normal-light images as can be seen in Fig.\ref{fig:Motivation}(c), are usually used as a strong LLIE-related constraint for the supervised/semi-supervised models to produce promising results \cite{LLNet,KinD,KinD++,DCC-Net,LLFlow,URetinex-Net,Retinexformer}. LLNet \cite{LLNet} is a pioneering deep LLIE method by training an auto-encoder to extract features and enhance images. KinD \cite{KinD} and KinD++ \cite{KinD++} are based on the retinex theory, which uses three deep sub-networks and obtains remarkable results compared to traditional retinex-based methods. LLFlow \cite{LLFlow} proposes a normalizing flow-based model for LLIE, addressing limitations of prior methods that relied on pixel-wise reconstruction losses and deterministic processes. By formulating the Retinex decomposition problem as an implicit priors regularized model, URetinex-Net \cite{URetinex-Net} successfully combines the strengths of model-based and learning-based methods for LLIE. Retinexformer \cite{Retinexformer} designs a novel Transformer-based model for LLIE and significantly outperforms existing methods, especially in removing noise and artifacts while preserving color. However, paired data are costly to be collected in reality, and the over-reliance on synthetic paired data will limit the practical applications due to the large distribution difference between the synthetic data and real data.

\textbf{Unpaired data}, including the non-corresponding low-light and low/normal-light images (see Fig.\ref{fig:Motivation}(d)), are used as an ordinary LLIE-related constraint to train unsupervised models, such as EnlightenGAN \cite{EnlightenGAN, PairLIE}. EnlightenGAN is based on the generative adversarial networks (GAN) \cite{GAN} and is the first to utilize fully unpaired data to train an unsupervised LLIE model. Unlike EnlightenGAN, PairLIE \cite{PairLIE} does not utilize non-corresponding low-light and normal-light images, but rather two low-light images to learn adaptive priors, which has yielded good LLIE effects. However, unsupervised methods usually rely on a huge number of parameters to compensate for the weak constraint caused by the unpaired data, which also hinders their practical applications for LLIE. 

\textbf{Zero-reference data}, i.e., single low-light image only (see Fig.\ref{fig:Motivation}(e)), facilitate a simple model that trains a DNN by an elaborate unsupervised weak constraint \cite{Zero-DCE, Zero-DCE++, RUAS, SCI}. For example, Zero-DCE \cite{Zero-DCE} and Zero-DCE++ \cite{Zero-DCE++} convert the LLIE task into deep curve estimation, and obtain visually pleasing results. RUAS \cite{RUAS} first introduces the neural architecture search (NAS) into LLIE task, and SCI \cite{SCI} develops a self-calibrated illumination learning framework with remarkably enhanced results by taking into account the computational efficiency, flexibility and robustness. These methods yield surprising results using only zero-reference data, and are far more lightweight and efficient than those methods using paired/unpaired data. However, their performances are still highly dependent on the training data, but it is hard to determine the most suitable data.

Overall, the above three types of data-driven methods each have their own advantages. For example, paired data-based methods can achieve very high performance on specific data, while zero-reference data-based methods usually have outstanding superiority in terms of training/inference speed and robustness. Unlike the above three types of methods, this paper offers another new alternative to complete the LLIE task, which possesses the characteristics of faster training and inference, smaller model size, suppresses overexposure, and easy integration with other tasks. Specifically, we introduce Noise SElf-Regression (NoiSER), by sampling the noise (see Fig.\ref{fig:Motivation}(f)) from a Gaussian distribution as both model input and supervised signal to learn the parameters, low-light images can be directly enhanced by the learned model. We summarize the main contributions as follows:
\begin{itemize}
	\vspace{-2mm}
	\item We introduce Noise SElf-Regression (NoiSER), a simple and effective method for the LLIE task. By training a simple self-regression model (SRM) using noise, the trained model can directly be used to enhance a low-light image. The biggest difference between NoiSER and other LLIE methods is that NoiSER uses pure Gaussian Noise as training data instead of those task-related data, e.g., paired/unpaired data and zero-reference data. 
	
	\item On several widely-used image datasets, our NoiSER is highly competitive with the other competitors that use different types of task-related data. In addition, NoiSER is lightweight and efficient in both training and inference phases. Specifically, with only about 1K parameters, NoiSER achieves about 1 minute for training and 1.2 milliseconds for inference on a 600$\times$400 resolution image by a single RTX 2080 Ti. As a bonus, NoiSER can also alleviate the issue of image overexposure to some extent.
	
	\item By meticulously designing a deep network, we have preliminarily proved that NoiSER can achieve joint processing with other tasks (e.g., deraining) without requiring LLIE data, indicating that NoiSER has great potential in joint processing.
\end{itemize}

\vspace{-1mm}
\section{Preliminaries}
\label{sec:preli}
\vspace{-0.5mm}
\subsection{Image Self-Regression Principle} 
\vspace{-0.5mm}
Image self-regression \cite{Auto-Encoder, VAE, Deep_Image_Prior} utilizes the input data itself as the supervised signal to reconstruct the output. By assuming $x$ follows a certain distribution, this process can be represented by minimizing the following empirical risk: 
\vspace{-1mm}
\begin{equation}\label{eqn1}
	\mathop{\arg\min}\limits_{\theta}\mathbb{E}_{x}\{L(f_\theta (x),x)\},
	\vspace{-1.5mm}
\end{equation}
where $f_\theta$ denotes a parametric family of mapping and $L$ is a loss function. After optimization, given an arbitrary input $x_0$, $f_\theta$ can yield a texture-similar output $x_0^\ast$. 

Recent years, image self-regression has been applied as a learning approach to a variety of tasks, e.g., dimension reduction \cite{Auto-Encoder}, image generation \cite{VAE}, and image restoration \cite{Deep_Image_Prior}, with significant and far-reaching impact. Deep image prior \cite{Deep_Image_Prior}, as the most relevant self-regression method, uses the corrupted image as input and supervised signal to reconstruct the natural image, which obtains stunning results in multiple tasks, such as denoising and super-resolution, implying that a DNN has an innate ability to learn the ``uncorrupted, natural" part of the image before learning the ``corrupted" part. In this work, we use self-regression to reconstruct the texture/contrast of low-light images. The designed self-regression model ($f_\theta$) is shown in Fig.\ref{fig:Architecture}. 

\vspace{-1mm}
\subsection{Gray-World Color Constancy Hypothesis}
\vspace{-0.5mm}
The gray-world color constancy hypothesis \cite{Gary-world_Hypothesis} (shortly, gray-world hypothesis) is useful in image processing, which says, for an image with large color variations, the averages of the three RGB components converge to the same gray value $K$. 
In recent years, some deep learning-based LLIE methods use the gray-world hypothesis as a loss function and obtain impressive LLIE performance \cite{Zero-DCE,Zero-DCE++,SCL_LLE}. Zero-DCE \cite{Zero-DCE} formulates the gray-world hypothesis as a color constancy constraint, which is expressed as
\vspace{-1mm}
\begin{equation}\label{eqn2}
	\small
	\mathcal{L}_{col}\!=\!\sqrt{\!\sum_{(p,q)\in S}(J^p\!-\!J^q)^2},S\!=\!\{(R,G),(R,B),(G,B)\},\!
\end{equation}
where $J^p$ and $J^q$ represent the average values of the $p$ and $q$ channels in the enhanced image, respectively.

\vspace{-1mm}
\section{Proposed Method}
\label{sec:propo}

\subsection{Motivation and Problem Statement}
Existing deep learning-based LLIE methods are mainly driven by three types of data, i.e., paired data, unpaired data, and zero-reference data. Obtaining these three types of data involves different costs, in descending order: paired data $>$ unpaired data $>$ zero-reference data. Additionally, determining the appropriate data to train the model can also be challenging. Therefore, we attempt to pose a question: Is it possible to use almost cost-free data, even task-irrelevant data, to train the model for the LLIE task? This would not only reduce the data cost but also avoid the problem of selecting training data.

If we can solve this problem, the proposed solution can serve as a new alternative for actual use by people. Additionally, considering that the solution has almost no data cost, we can also speculate a new usage of the solution, namely to combine it with other tasks for joint processing without introducing additional LLIE data, similar to some existing joint processing methods, e.g., joint LLIE tasks with deraining \cite{Low_Light_Deraining}, denoising \cite{Low_Light_Denoising, Low_Light_Denoising_and_Deblurring}, object/face detection \cite{Low_Light_Face_Detection, Low_Light_Face_Detection_1, Low_Light_Face_Detection_2,Low_Light_Object_Detection} or semantic segmentation \cite{Low_Light_Semantic_Segmentation}.

\begin{figure}[t]
	\centering
	\includegraphics[width=1\columnwidth]{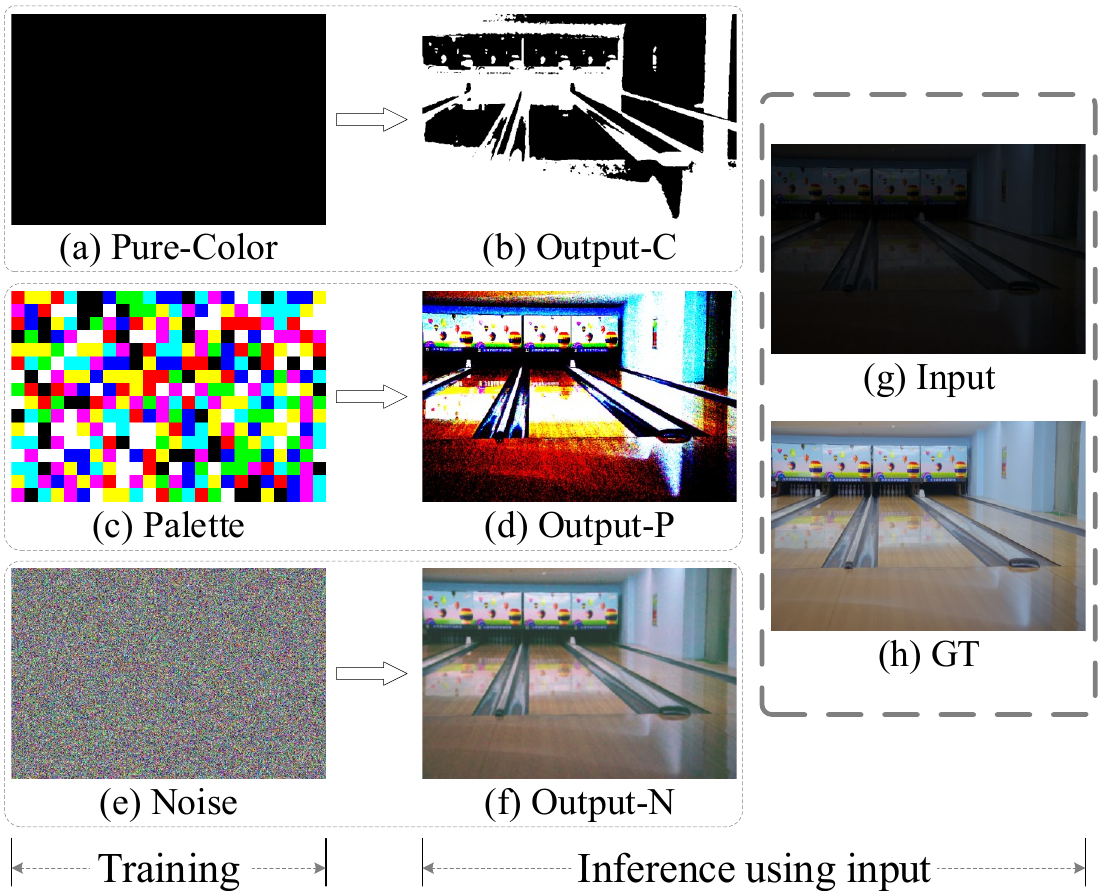}
	\vspace{-4mm}
	\caption{Comparison of the fully-converged enhancement results using different training data. Clearly, the noise self-regression approach can yield visually better results.}
	\label{fig:SRDataResult}
	\vspace{-4mm}
\end{figure}

\vspace{-1mm}
\subsection{Solution}
\vspace{-0.5mm}
\subsubsection{Abstracting and Meeting Task Requirements}\label{subsubsec:requirments}
\vspace{-0.5mm}
Let's rewrite the problem, i.e., ``can we enhance a low-light image by deep learning without any task-related data?"
We first decompose and abstract it into three requirements: 1) any task-related data cannot be used for training; 2) given an image, the output of the model should have similar adjacent pixel contrast (texture) to the input; 3) given a low-light image, the output of the model should be of normal light.

For the requirement 1), the easiest solution is to use the random noise to replace the task-related data. Disregarding how to use the random noise, at least we have data for training. The requirement 2) gives us a message that we need to train a model with a reconstruction ability. Considering the image self-regression principle, we can use the noise itself as supervisory signals to train the model. Now, we can satisfy the requirements 1) and 2) in a self-regression manner by using random noise. For the requirement 3), we can force the output to satisfy the $K$-value gray-world hypothesis, where $K$ should be in a normal gray range, such as 80-140. The proof about why this manner could convert light to normal is given in \textit{Proposition 3} of Section \ref{subsubsec:P-regression}.

We show the ultimate pipeline of training and inference, and the detailed structure of our self-regression model for an intuitive observation in Fig.\ref{fig:Architecture}. It is noteworthy that the noise itself is still relatively complex data. As such, prior to training a model with noise, we introduce two types of data with much simpler structures, i.e., pure-color (see Fig.\ref{fig:SRDataResult}(a)) and palette (see Fig.\ref{fig:SRDataResult}(c)). In what follows, we introduce three new self-regressions asymptotically.

\vspace{-1mm}
\subsubsection{Pure-\underline{C}olor Self-\underline{Regression} (C-Regression)}\label{subsubsec:C-regression}
\vspace{-0.5mm}

\begin{figure}[t]
	\centering
	\includegraphics[width=1\columnwidth]{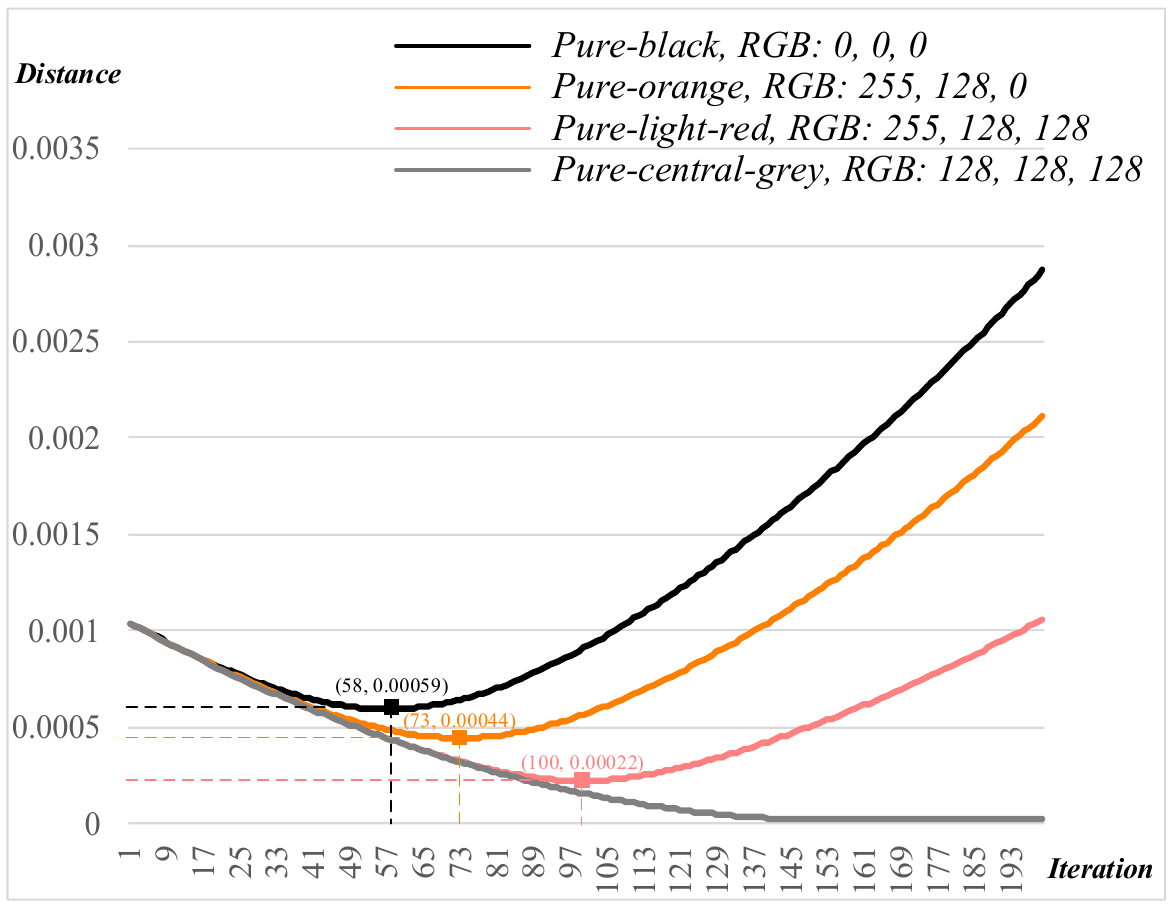}
	\vspace{-4mm}
	\caption{Iteration-Distance curves using different pure-colors for C-regression training. The distance is the $L_2$ norm between the output and the pure-central-grey during training. The pure-central-grey curve did not converge to $0$ but to a very small number.}
	\label{fig:Proposition1}
	\vspace{-3mm}
\end{figure}

Considering the general $256$ grey level, we first define the pure-color images (shortly, pure-color) as $I_c$ $\in$ $\mathbb{N}^{H\times W\times 3}$ under the following constraints:
\vspace{-1mm}
\begin{equation}\label{eqn3}
	\begin{aligned}
		& \forall p_i,p_j\in I_c,\quad p_i=p_j\\
		& \forall p_i\in I_c,\quad max(p_i)<=255,
	\end{aligned}
\end{equation}
where $p_i,$$p_j$ denote pixels with three channels and $max(\cdot)$ is maximum operation. Term ``pure-" denotes an image rather than a color, e.g., pure-black and pure-light-red images.

\begin{figure*}[t]
	\centering
	\includegraphics[width=2\columnwidth]{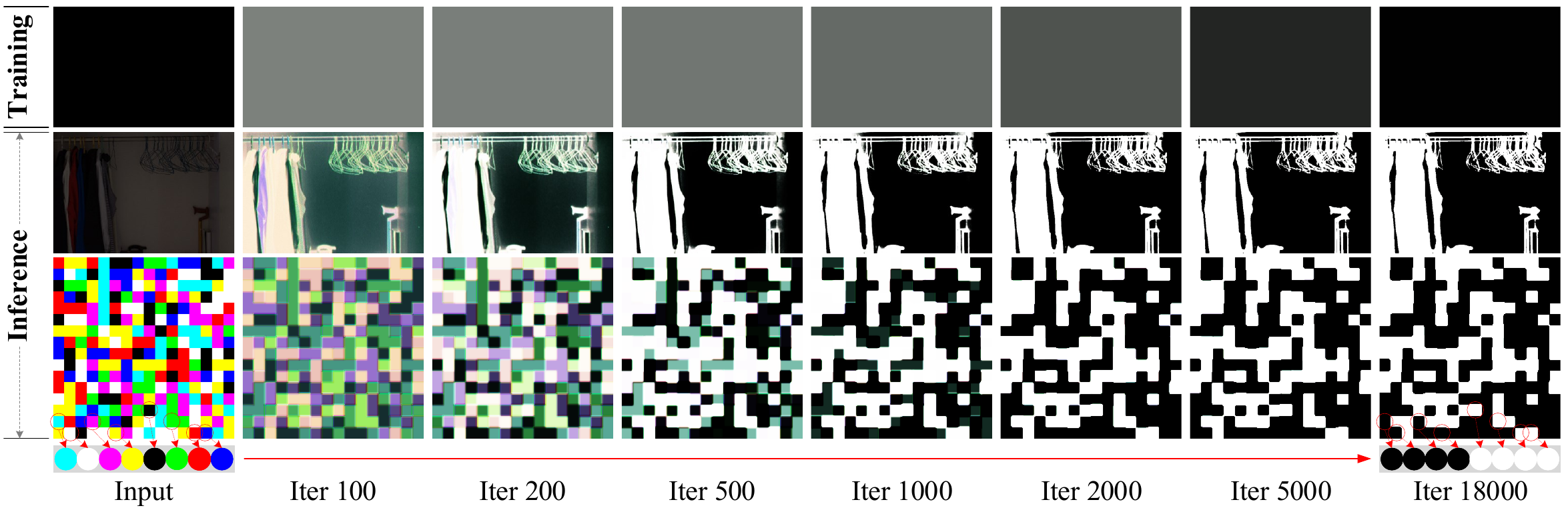}
	\vspace{-2mm}
	\caption{The phenomenon obtained from the pure-black self-regression. For an arbitrary input with a degree of contrast, the model finally tend to divide the input to be either black or white.}
	\vspace{-2mm}
	\label{fig:PureBlackSR}
\end{figure*}

We first use pure-color for self-regression training due to two main reasons: 1) some low-light images themselves can be seen as pure-black with a degree of contrast, and 2) pure-color is very simple, which makes it easier to understand the effect of C-regression in enhancing low-light images, which is also instructive for the use of other more complex data.

Following the general image self-regression principle, C-regression minimizes the deviation between model output and the pure-color $I_c$ itself according to a loss function: 
\vspace{-1mm}
\begin{equation}\label{eqn4}
	\mathop{\arg\min}\limits_{\theta}\mathbb{E}_{I_c}\{L(f_\theta (I_c),I_c)\},
\end{equation}
where $f_\theta$ is a parametric family of mappings and $L_1$ is used to minimize the reconstruction error over pure-color. 

We first take the pure-black as an example for training and directly apply the trained model to enhance low-light image, as shown in Fig.\ref{fig:SRDataResult}(g) and Fig.\ref{fig:SRDataResult}(b). We can easily see that the low-light input is crudely broken down into binary colors, i.e., black and white, which implies that pure-black cannot enhance image although it is close to a dark image. However, we can still gain some inspiration, e.g., C-regression has roughly reconstructed the texture of the input image. Therefore, it will be crucial to reveal the mechanism on how the input is divided into two colors. 

Deep neural networks often work in a black box manner and we cannot explain the intrinsic mechanism at the micro level. But at the macro level, we can still obtain a correct conclusion based on induction reasoning\cite{InductionReasoning1,InductionReasoning2}. Specifically, we follow the steps of incomplete induction reasoning, i.e., \textit{individual phenomenon} $\rightarrow$ \textit{individual conclusion} $\rightarrow$ \textit{universal conclusion} $\rightarrow$ \textit{intra-domain validation}. Specifically, based on the phenomenon of pure-black self-regression (i.e., dividing low-light input into black and white), we can induce an individual conclusion and then generalize it from individual to universal, and as a result, this universal conclusion can be applied to other pure-color (e.g., pure-red). Then, we need to perform self-regression training using other colors, and if the results also satisfy the universal conclusion, this conclusion can be proven to be correct. Prior to performing the incomplete induction reasoning, we introduce the required definitions and propositions.

\vspace{1mm}
\textit{Definition 1 (central grey)}: For a specific range of image values [a,b], the central grey is a color $C_g\in \mathbb{R}^3$  if it satisfies
\vspace{-2mm}
\begin{equation}\label{eqn5}
	C_g^i=\frac{a+b}{2}, \quad i\in \{R,G,B\},
\end{equation}
where $C_g$ $=$ $(127.5,127.5,127.5)$ in the general 256 grey level (range: [0, 255]), while actually the channel value is replaced by 128, since only integers are allowed.
\begin{table*}[t]
	\footnotesize
	\centering
	\vspace{1mm}
	\caption{Illustration of the C-regression mechanism.}
	\vspace{-2mm}
	\setlength{\tabcolsep}{0.3mm}{
		\begin{tabular}{c|c|c|cccccccc|c|c|c|cccccccc}
			\hline
			\multicolumn{11}{c|}{\textbf{(a) Individual phenomenon $\rightarrow$ Individual conclusion}} & \multicolumn{11}{|c}{\textbf{(b) Universal conclusion $\rightarrow$ Intra-domain validation}}\\
			\hline
			\hline
			\multirow{3}{*}{\tabincell{c}{Training \\(Black)}} & \color{green}{\textbf{1}} & Initial & R:128 & G:128 & B:128 & \multicolumn{5}{|c|}{ } & \multirow{3}{*}{\tabincell{c}{Training \\(Red)}} & \color{red}{\textbf{1}} & Initial & R:128 & G:128 & B:128 & \multicolumn{5}{|c}{ } \\
			& \color{green}{\textbf{2}} & Final & R:0 & G:0 & B:0 & \multicolumn{5}{|c|}{Black or White?} & & \color{red}{\textbf{2}} & Final & R:255 & G:0 & B:0 & \multicolumn{5}{|c}{Red or Cyan?} \\
			& \color{green}{\textbf{3}} & Trend & R:$\downarrow$ & G:$\downarrow$ & B:$\downarrow$ & \multicolumn{5}{|c|}{ } & & \color{red}{\textbf{3}} & Trend & R:$\uparrow$ & G:$\downarrow$ & B:$\downarrow$ & \multicolumn{5}{|c}{ } \\
			\hline
			\multirow{9}{*}{Inference} & \color{blue}{\textbf{4}} & Color & White & Cyan & Purple & Yellow & Red & Green & Blue & Black & \multirow{9}{*}{Inference} & \color{red}{\textbf{4}} & Color & White & Cyan & Purple & Yellow & Red & Green & Blue & Black \\
			
			& \color{green}{\textbf{5}} & R$\downarrow$ & 255 & 0 & 255 & 255 & 255 & 0 & 0 & 0  & & \color{red}{\textbf{5}} & R$\uparrow$ & 255 & 0 & 255 & 255 & 255 & 0 & 0 & 0 \\
			
			& \color{green}{\textbf{6}} & G$\downarrow$ & 255 & 255 & 0 & 255 & 0 & 255 & 0 & 0 & & \color{red}{\textbf{6}} & G$\downarrow$ & 255 & 255 & 0 & 255 & 0 & 255 & 0 & 0 \\
			& \color{green}{\textbf{7}} & B$\downarrow$ & 255 & 255 & 255 & 0 & 0 & 0 & 255 & 0 & & \color{red}{\textbf{7}} & B$\downarrow$ & 255 & 255 & 255 & 0 & 0 & 0 & 255 & 0 \\
			& \color{green}{\textbf{8}} & R?& \checkmark &  & \checkmark & \checkmark & \checkmark &  &  && & \color{red}{\textbf{8}} & R? &  & \checkmark &  &  &  & \checkmark & \checkmark & \checkmark \\
			& \color{green}{\textbf{9}} & G? & \checkmark & \checkmark &  & \checkmark &  & \checkmark &  &  & & \color{red}{\textbf{9}} & G? & \checkmark & \checkmark &  & \checkmark &  & \checkmark &  &  \\
			& \color{green}{\textbf{10}} & B? & \checkmark & \checkmark & \checkmark &  &  &  & \checkmark & & & \color{red}{\textbf{10}} & B? & \checkmark & \checkmark & \checkmark &  &  &  & \checkmark &  \\
			& \color{green}{\textbf{11}} & No. & 3 & 2 & 2 & 2 & 1 & 1 & 1 & 0 & & \color{red}{\textbf{11}} & No. & 2 & 3 & 1 & 1 & 0 & 2 & 2 & 1 \\
			
			& \color{blue}{\textbf{12}} & To & Black & Black & Black & Black & White & White & White & White & & \color{red}{\textbf{12}} & To & \color{red}{Red\textbf{?}} & \color{red}{Red\textbf{?}} & \color{red}{Cyan\textbf{?}} & \color{red}{Cyan\textbf{?}} & \color{red}{Cyan\textbf{?}} & \color{red}{Red\textbf{?}} & \color{red}{Red\textbf{?}} & \color{red}{Cyan\textbf{?}}\\
			\hline
	\end{tabular}}
	\label{tab1}
	\vspace{-5mm}
\end{table*}

\vspace{1mm}
\textit{Definition 2 (opposite color)}: For a specific range of image values [a,b], a color $C_1\in \mathbb{R}^3$ is the opposite color of the color $C_2\in \mathbb{R}^3$ if they satisfy the following condition: 
\begin{equation}\label{eqn6}
	\lvert C_1^i-C_g^i\rvert=\lvert C_2^i-C_g^i\rvert, \quad i\in \{R,G,B\},
\end{equation}
where $C_g$ denotes the central grey, and $C_1$$\neq$$C_2$.

\vspace{1mm}
\textit{Proposition 1:} Based on a model with random initialization, arbitrary C-regression tends to construct the pure-central-grey in the initial iterations.

\vspace{1mm}
\textit{Validation for Proposition 1:} Given an image $I$, we use Eqn.\ref{eqn2} to measure the distance between this image and the pure-central-grey with small modification as follows: 
\vspace{-1mm}
\begin{equation}\label{eqn7}
	\mathcal{D}(I)=\sqrt{\sum_{i\in S}(I^i-C_g^i)^2},\quad S=\{R,G,B\},
	\vspace{-1mm}
\end{equation}
where $I^i$ is the average value of the $i$ channel in image $I$.

In this study, we choose four pure-colors with different distances from the pure-central-grey for validation, i.e., pure-black (RGB(0,0,0)), pure-orange (RGB(255,128,0)), pure-light-red (RGB(255,128,128)) and the pure-central-grey itself (RGB(128,128,128)). Clearly, the relation of the distances satisfies: $\mathcal{D}(\operatorname{pure-black})$ $>$ $\mathcal{D}(\operatorname{pure-orange})$ $>$ $\mathcal{D}(\operatorname{pure-light-red})$ $>$ $\mathcal{D}(\operatorname{pure-central-grey})$ $=$ $0$. In Fig.\ref{fig:Proposition1}, we show the iteration-distance curves based on these pure-colors during C-regression training. For non-pure-central-grey self-regression training, no supervised signals are pure-central-grey; nevertheless, the overall trend of the curves is initially down and then up rather than directly up, which indicates \textit{Proposition 1} is True. For the pure-grey self-regression training, the curve falls and converges directly, which additionally proves the \textit{Proposition 1}.

\textbf{\textit{Individual phenomenon.}} Fig.\ref{fig:SRDataResult}(a), Fig.\ref{fig:SRDataResult}(g) and Fig.\ref{fig:SRDataResult}(b) show an example of using pure-black self-regression training to infer a low-light image. We see from Fig.\ref{fig:SRDataResult}(h) and Fig.\ref{fig:SRDataResult}(b) that the trained model tends to map some colors (e.g., white and yellow) to the color used in C-regression training (i.e., black, RGB(0,0,0)), while mapping the others (e.g., black, red and green) to the opposite color (i.e., white, RGB(255,255,255)). This inspires us to explore the mapping relationship of the colors before and after inference. To this end, we build a palette containing rich colors as the image to be inferred. Fig.\ref{fig:PureBlackSR} shows the process of inferring a low-light image and shows a palette in pure-black self-regression iterative training. We see that there are 8 colors in the palette, of which 4 colors (i.e., cyan, white, purple and yellow) are mapped to black and the other 4 colors (i.e., black, green, red and blue) are mapped to the opposite color of black, which is applicable to each pixel in the palette. 

\begin{figure}[t]
	\centering
	\includegraphics[width=1\columnwidth]{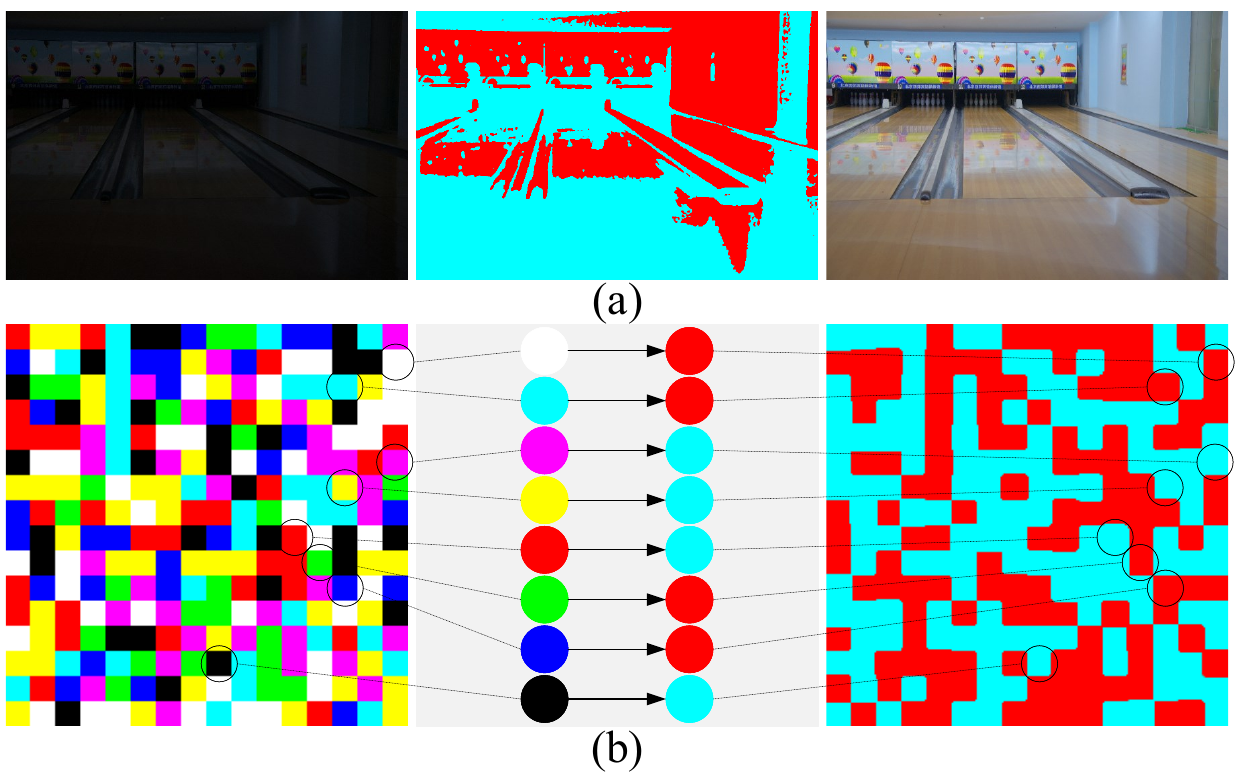}
	\vspace{-6mm}
	\caption{Intra-domain validation via pure-red self-regression.}
	\label{fig:CRValidation}
	\vspace{-4mm}
\end{figure}

\textbf{\textit{Individual conclusion.}} Table \ref{tab1}(a) illustrates the specific induction process from individual phenomenon to individual conclusion based on pure-black self-regression. Specifically, the individual phenomenon is shown in rows 4 and 12, while the induction process of the individual conclusion is shown in rows 1-3 and 5-11. During training, \textit{Proposition 1} tells us that the initial output tends to the pure-central-grey (row 1); as the iterations increase, the output becomes very close to the pure-black (row 2). Thus, we can find that for all three RGB channels, the pure-black self-regression perhaps learns a downtrend (row 3). During inference, we see from the phenomenon that the input is divided into black and white (rows 4 and 12), which may be related to the learned trends. To this end, firstly, we list the RGB values (rows 5-7) of the colors in the fourth row; secondly, we check whether the specific channel satisfies the learned trend (rows 8-10). Taking ``Cyan" as an example, since all three RGB channels learn to trend down, the G,B channels in ``Cyan" satisfies the trend, but R channel does not (0 will overflow if a downward trend is taken); thirdly, we record the number of channels that satisfy the learned trend (row 11); finally, based on the results in row 11 and the observations in row 12, we believe if a color meets at least two channel trends, it will be mapped to black (i.e., training color), and conversely, to white (the opposite of training color).

\textbf{\textit{Universal conclusion.}} According to the individual conclusion, we can directly generalize it to a universal one, i.e., \textit{during training, C-regression attempts to learn the trends of the three channels of RGB. During inference, if a color satisfies over two channel trends, it will be mapped to the training color, otherwise, to the opposite of training color.}

\textbf{\textit{Intra-domain validation.}} To verify the correctness of the universal conclusion, we need to perform C-regression training based on the other colors and see whether the experimental results are consistent with this conclusion. Taking the pure-red self-regression as an example, we perform the reasoning process in Table \ref{tab1}(b). Row 12 displays the reasoning results, from which we see that white, cyan, green and blue are mapped to red, while purple, yellow, red and black are mapped to cyan. We further conduct and display the corresponding experiments in Fig.\ref{fig:CRValidation}. Clearly, the results keep consistent with the reasoning results, which proves that the conclusion is correct. Here, there is one point that must be noticed, i.e., the condition for the above conclusion to be valid is that the distribution between the training and inference data is far different.

At this point, we have fully understood the mechanism of C-regression and can assert the following facts:
\begin{itemize}
	\vspace{-2mm}
	\item C-regression preliminarily achieved texture reconstruction but cannot reconstruct details, i.e., C-regression has already met requirement (1) in Section \ref{subsubsec:requirments} but has not fully met requirement (2).
	\item The color binarisation of the C-regression results relies on the color homogeneity in the training samples, and it is conceivable that a more accurate texture may be reconstructed if each training sample is enriched with multiple colors and has a degree of contrast.
\end{itemize}

\begin{table*}[t]
	\centering
	\caption{List of RGB channel means for different datasets under the general $256$ grey level. We see that for an image of normal light, the RGB channel means usually fall within [80, 140]. The second column is the mean of the noise sampled by $\mathcal{N}(0,\sigma^2)$ (0 for [-1,1] is equivalent to 128 for [0,255]). The term ``Total" indicates that all images of LOL, LSRW (Huawei) and LSRW (Nikon) datasets are used.}
	\vspace{-2mm}
	\setlength{\tabcolsep}{1.2mm}{
		\begin{tabular}{c|c|c|cc|cc|cc|ccc}
			\hline
			Sets & Noise & Total & \multicolumn{2}{c}{\textbf{LOL \cite{RetinexNet}}} & \multicolumn{2}{|c|}{\textbf{\tabincell{c}{LSRW (Huawei) \cite{Dataset-LSRW}}}} & \multicolumn{2}{|c|}{\textbf{\tabincell{c}{LSRW (Nikon) \cite{Dataset-LSRW}}}} & \multicolumn{3}{c}{\textbf{SICE \cite{Dataset-SICE}}}\\
			\hline
			\tabincell{c}{Light} & - & Normal & \tabincell{c}{Normal} & \tabincell{c}{Low} & \tabincell{c}{Normal} & \tabincell{c}{Low} & \tabincell{c}{Normal} & \tabincell{c}{Low}  & \tabincell{c}{Normal} & \tabincell{c}{Overdark} & \tabincell{c}{Overexposed} \\
			\hline
			R & 128 & 116.50 & 120.47 & 15.49 & 110.39 & 18.92 & 122.69 & 42.26 & 122.12 & 32.78 & 164.82 \\
			G & 128 & 109.93 & 113.88 & 15.22 & 98.22 & 16.39 & 124.55 & 43.19 & 121.18 & 32.71 & 161.40 \\
			B & 128 & 103.82 & 110.38 & 14.73 & 86.63 & 15.20 & 124.69 & 43.48 & 109.77 & 31.07 & 149.51 \\
			\hline
	\end{tabular}}
	\label{tab2}
	\vspace{-4mm}
\end{table*}

\vspace{-4mm}
\subsubsection{\underline{P}alette Self-\underline{Regression} (P-Regression)}\label{subsubsec:P-regression}
\vspace{-0.5mm}
In Section \ref{subsubsec:C-regression}, we have used the palette for inference, and here we will illustrate that using the palette as training data for self-regression allows for a generation of normal-light image and a finer reconstruction of the image textures. Considering the general $256$ grey level, we can define the palette as $I_p$ $\in$ $\mathbb{N}^{H\times W\times 3}$ with the following constraints:
\vspace{-1mm}
\begin{equation}\label{eqn8}
	\begin{aligned}
		& \forall q_i,q_j\in I_p,\quad size(q_i)=size(q_j)\\
		& \forall q_i \in I_p, \;\forall p_i, p_j\in q_i,\quad p_i=p_j\\
		& \forall q_i \in I_p, \;\forall p_i \in q_i, \quad max(p_i) <=255,
	\end{aligned}
	\vspace{-1mm}
\end{equation}
where $q_i$,$q_j$ denote non-overlapping patches, $size(\cdot)$ denotes the grain/size of the patch (setting size to 16$\times$16 in this work), $p_i,$$p_j$ denote pixels with three channels and $max(\cdot)$ denotes the maximum operation. From the above definition, the palette can be regarded as the coarse-grained noise and as the fine-grained pure-color.

\begin{figure}[t]
	\centering
	\includegraphics[width=1\columnwidth]{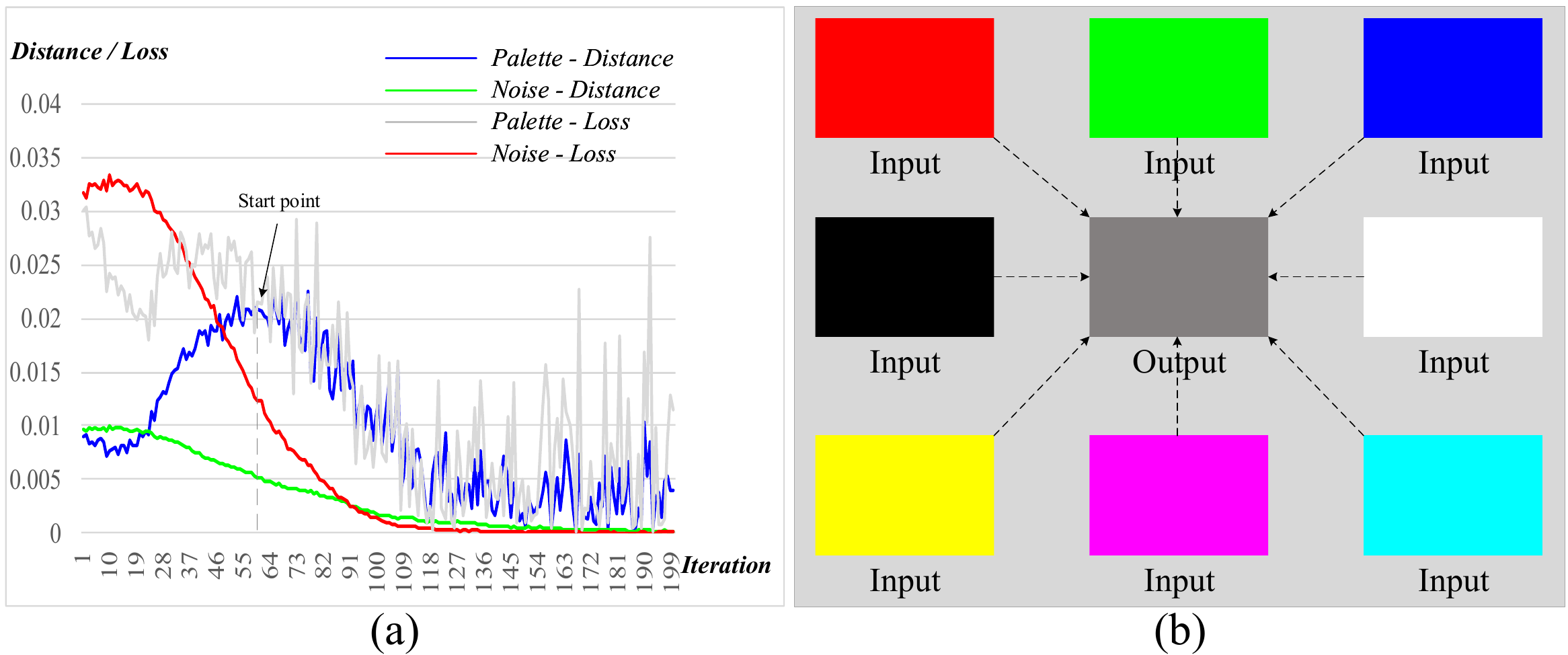}
	\vspace{-4mm}
	\caption{(a) Iteration-Distance/Loss curves for P-regression and NoiSER training. The distance is $L_2$ norm between output and pure-central-grey during training, and the loss is color constancy loss of output. (b) Pure-color mappings after training.}
	\label{fig:Proposition24}
	\vspace{-2mm}
\end{figure}

Similar to C-regression, given palettes $I_p$ as training data, P-regression is the process whereby the palette learns to rebuild itself according to certain loss function:
\vspace{-2mm}
\begin{equation}\label{eqn9}
	\mathop{\arg\min}\limits_{\theta}\mathbb{E}_{I_p}\{L(f_\theta (I_p),I_p)\},
	\vspace{-2mm}
\end{equation}
where we also employ the $L_1$ loss for minimizing the reconstruction error based on the palette.

Fig.\ref{fig:SRDataResult}(g) and Fig.\ref{fig:SRDataResult}(d) display an example of inferring a low-light image through P-regression. It is clear that P-regression successfully reconstructs the textures of the input low-light image. Next, we introduce the required propositions to figure out the mechanism behind P-regression. 

\vspace{0.5mm}
\textit{Proposition 2:} Based on a model with random initialization, P-regression tends to start building the pure-central-grey at a starting point and continues until it converges.

\vspace{0.5mm}
\textit{Validation for Proposition 2:} Given an image $I$, similar to C-regression, we use Eqn.\ref{eqn7} to measure the distance between this image and the pure-central-grey. The blue curve in Fig.\ref{fig:Proposition24}(a) shows the distance between the output and pure-central-grey. We see that after a certain starting point (about the 61st iteration), the output gradually approaches the pure-central-grey, although we did not add any constraints during training, which means that \textit{Proposition 2} is True. Besides, we use pure-colors as inference images for validation, and obtain a consistent conclusion with \textit{Proposition 2}, i.e., all outputs appear grey (see Fig.\ref{fig:Proposition24}(b)).

\vspace{0.5mm}
\textit{Proposition 3:} Known \ding{172}: The image satisfies the $K$-value gray-world hypothesis, where $K$ is approximately between 80 and 140; Known \ding{173}: The image is of normal light. Then, \ding{172} is a statistically necessary condition for \ding{173}.

\vspace{0.5mm}
\textit{Validation for Proposition 3:} In Table \ref{tab2}, we describe the RGB channel means (i.e., mean value of each color channel) over several widely-used-datasets, LOL \cite{RetinexNet}, LSRW \cite{Dataset-LSRW} and SICE \cite{Dataset-SICE}. We conclude that: (1) all datasets satisfy the gray-world hypothesis (the channel deviation is also not too large for LSRW (Huawei)-Normal); (2) the channel means of the normal datasets lie between 80-140. Note that (1) and (2) provide the proof of \textit{Proposition 3}.

For P-regression, the training samples themselves have a degree of contrast (i.e., variation between non-overlapping patches), and based on the self-regression principle, the contrast can be reconstructed after self-regression training, which enables it to generate the texture and structure of the input image more accurately than C-regression. In addition, according to the \textit{Propositions 2} and \textit{3}, we know the P-regression has the ability to map any color to approach grey, which means the channel means of the output meet the K range (80-140) of the normal-light images. As such, suppose the P-regression further satisfies the gray-world hypothesis and we can use P-regression to get the normal-light output that satisfies all three requirements mentioned in Section \ref{subsubsec:requirments}. However, this supposition is valid but not strictly. We use the color constancy loss (Eqn.\ref{eqn2}) to measure the satisfaction degree towards the gray-world hypothesis (not applying it to back propagation). As the grey curves in Fig.\ref{fig:Proposition24}(a) show, the overall downward trend suffers from sharp fluctuations. Nevertheless, the P-regression can still yield a relatively favorable result.

Now, we can clarify the mechanism of P-regression and assert the following facts:
\begin{itemize}
	\vspace{-2mm}
	\item P-regression initially meets all the three requirements mentioned above, which means it can already enhance a low-light image to some extent. 
	\item The enhancement results of P-regression are unpleasant due to the sharp fluctuations in convergence to gray-world hypothesis. It is conceivable that a more pleasing result might be generated, if we can smooth out the sharp curve fluctuations.
\end{itemize}

\begin{figure}[t]
	\centering
	\includegraphics[width=0.98\columnwidth]{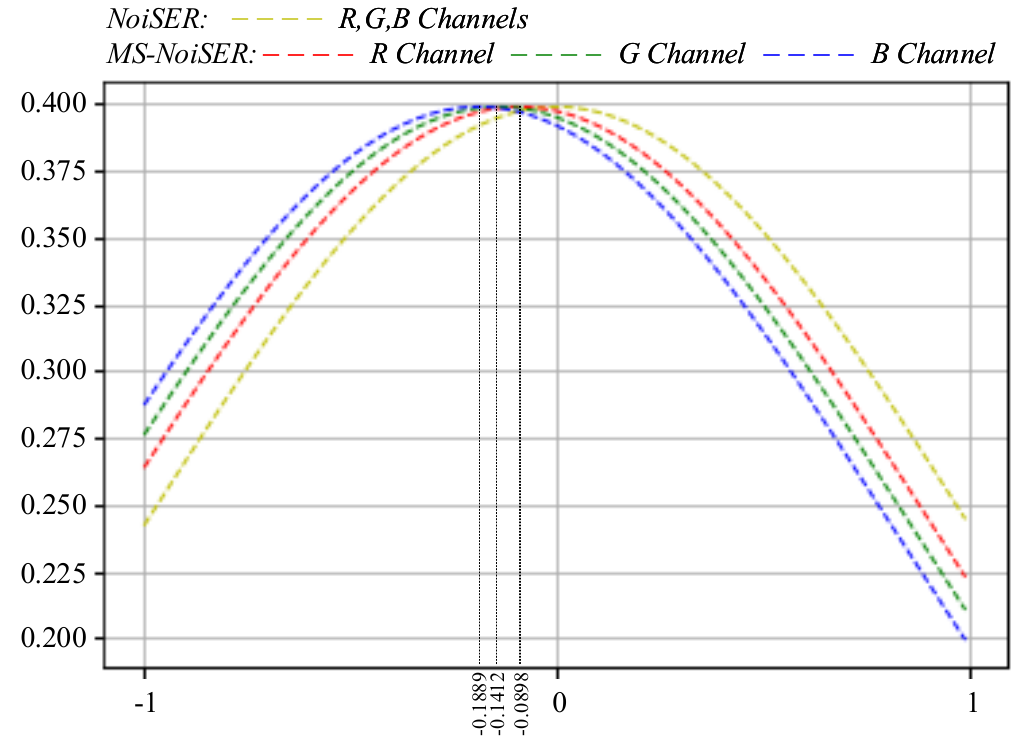}
	\vspace{-2mm}
	\caption{The difference of the used distributions between NoiSER and MS-NoiSER. Comparing to the zero-mean Gaussian distribution used in NoiSER (shown in yellow dotted line), the means of the Gaussian distribution used in MS-NoiSER have shifts of the difference degrees on RGB channels (red/green/blue dotted line).}
	\label{fig:Mean-Shifted_Gaussian}
	\vspace{-4mm}
\end{figure}

\vspace{-4mm}
\subsubsection{\underline{Noi}se \underline{Se}lf-\underline{R}egression (NoiSER)}\label{subsubsec:NoiSER}
\vspace{-0.5mm}
Considering that one possible reason for sharp curve fluctuations is the larger grain size of the patches in a palette, we attempt to use noise for self-regression training, since noise can be approximated as a fine-grained palette.

Instead of defining an additional noise, we directly sample a Gaussian noise $I_n\sim \mathcal{N}(0,\sigma^2)$ for each pixel, and the specific regression process can be expressed as follows:
\vspace{-1mm}
\begin{equation}\label{eqn10}
	\mathop{\arg\min}\limits_{\theta}\mathbb{E}_{I_n}\{L(f_\theta (I_n),I_n)\},
	\vspace{-1mm}
\end{equation}
where $L_1$ loss is used to minimize the reconstruction error. 

Fig.\ref{fig:SRDataResult}(g) and Fig.\ref{fig:SRDataResult}(f) show an example of inferring a low-light image via NoiSER. As can be seen, given a low-light input, NoiSER successfully generates a visually pleasing enhanced result. We first introduce a required proposition to understand the specific mechanisms of NoiSER.

\vspace{0.5mm}
\textit{Proposition 4:} Based on a model with random initialization, NoiSER tends to approach pure-central-grey while satisfying the gray-world hypothesis.

\vspace{0.5mm}
\textit{Validation for Proposition 4:} Given an image $I$, similar to P-regression, we use Eqn.\ref{eqn7} to measure the distance between $I$ and the pure-central-grey, and use Eqn.\ref{eqn2} to measure the satisfaction degree towards the gray-world hypothesis. The green and red curves in Fig.\ref{fig:Proposition24}(a) demonstrate the specific iterative processes. As can be observed, although we do not apply the equation for back propagation, the curves still converge smoothly, which demonstrates the correctness of \textit{Proposition 4}. Similar to P-regression, NoiSER training can also make the pure-colors appear grey (see Fig.\ref{fig:Proposition24}(b)). We summarize the \textit{Proposition 4} as the fact that NoiSER has the ability to learn a gray-world mapping.

It is noted that NoiSER satisfies all the properties of P-regression while smoothing the convergence curve towards the gray-world hypothesis, which means that NoiSER has possessed the ability to enhance low-light image with pleasing visual effect. The pipeline of training and inference for NoiSER are shown in Fig.\ref{fig:Architecture}, where the designed SRM is equipped with a instance-normalization layer. Besides, the inputs/outputs of the model are normalized into the range [-1,1], the median of which is exactly the mean $\mu$ of the standard Gaussian distribution, i.e., $\mu=0$. In such setting, 0 roughly corresponds to the means of RGBs for normal-light image (see Table \ref{tab2}), i.e., zero-mean Gaussian distribution roughly describes the distribution of normal-light image.

Through the above, we can conclude that the IN layer may naturally remediate the overall magnitude/lighting of the input image. In addition, according to the image self-regression principle, the noise self-regression itself can reconstruct the texture and maintain a similar contrast between adjacent pixels as the input image. Combining these two, low-light image can be enhanced properly. However, if the output doesn't meet the gray-world hypothesis, visual effect will become poor (e.g., P-regression in Section \ref{subsubsec:P-regression}). Fortunately, Fig.\ref{fig:Proposition24} demonstrates that NoiSER satisfies the gray-world hypothesis, which means that NoiSER may enhance low-light images with visually pleasing effects.

\vspace{-1mm}
\subsubsection{\underline{M}ean-\underline{S}hifted \underline{NoiSER} (MS-NoiSER)}
\label{subsubsec:MS-NoiSER}
\vspace{-0.5mm}
We have shown that the self-regression for the noise sampled from Gaussian distribution $\mathcal{N}(0,\sigma^2)$ provides surprising enhanced results, in which the key point is that zero-mean Gaussian distribution roughly describes the distribution of normal-light image. However, $0$ may not be precise enough for simulating the mean of normal-light image in reality, although it can work effectively. Therefore, we also present a Mean-Shifted NoiSER (MS-NoiSER) strategy, which uses the mean-shifted Gaussian noise for self-regression. Specifically, given the means on the RGB channels for a specific dataset, we sample noise from the Gaussian distribution $\mathcal{N}(\mu, \sigma^2)$ on each channel independently for self-regression, in which the mean $\mu$ for the Gaussian distribution on each channel has different shift degrees compared to the zero-mean Gaussian distribution $\mathcal{N}(0, \sigma^2)$ and satisfies the given means of a specific dataset. For example, we can make the means on the RGB channels of Gaussian distribution meet the channel means of multiple datasets (see the third column in Table \ref{tab2}), based on which we use Fig.\ref{fig:Mean-Shifted_Gaussian} to show the difference of the used distributions between NoiSER and MS-NoiSER. The channel means $116.50/109.93/103.82$ in range [0,255] are denoted by $-\!0.0898/-\!0.1412/-\!0.1889$ in range [-1,1]. We show the superior performance of MS-NoiSER in Section \ref{sec:exper}.

\textbf{Remark.} Strictly speaking, MS-NoiSER is not a real sense method to learn a ``general generalization" ability, as the self-regression training needs the channel means of the dataset as prior. On the other hand, MS-NoiSER is also not a strict fitting method as current approaches, since the images in the dataset are not directly used for training. In practice, we can use NoiSER directly to learn a general generalization for LLIE. However, if NoiSER does not perform well on a certain dataset, e.g., LSRW (Huawei), since it's channel means is far from 128 (see the 6th column in Table \ref{tab2}), in such case we can optimize the result by MS-NoiSER.

\begin{table*}[t]
	\centering
	\caption{Numerical results on LOL dataset \cite{RetinexNet}, with best performance marked in red and the second best marked in blue. Clearly, our NoiSER has significant advantages, not only for performance metrics but also for application metrics.}
	\vspace{-2mm}
	\setlength{\tabcolsep}{1.6mm}{
		\begin{tabular}{c|c|c|ccc|ccc}
			\hline
			& \multirow{2}{*}{Training data} & \multirow{2}{*}{Methods} & \multicolumn{3}{|c|}{Performance metrics} & \multicolumn{3}{c}{Application metrics} \\
			\cline{4-9}
			&  &  & PSNR$\uparrow$ & SSIM$\uparrow$ & NIQE$\downarrow$ & TT$\downarrow$(\textit{min}) & IT$\downarrow$(\textit{ms}) & No.P$\downarrow$ \\
			\hline
			\hline
			\multirow{2}{*}{\tabincell{c}{Optimization\\-based}} & \multirow{2}{*}{-} & LIME \cite{LIME} & 14.2216 & 0.5144 & 8.5828 & - & 104553.82 & - \\
			&  & Zhang et al. \cite{DUAL} & 14.0181 & 0.5130 & 8.6111 & - & 138127.48 & - \\
			\hline
			\multirow{9}{*}{\tabincell{c}{Deep learning\\-based}} & Paired & RetinexNet \cite{RetinexNet} & 16.7740 & 0.4191 & 9.7294 & 3.17 & 95.40 & 1,333,841 \\
			\cline{2-9}
			& Unpaired & EnlightenGAN \cite{EnlightenGAN} & {\color{red}18.5413} & {\color{blue}0.6880} & 5.7111 & 90.05 & 10.91 & 6,959,553 \\
			\cline{2-9}
			& \multirow{4}{*}{Zero-reference} & Zero-DCE \cite{Zero-DCE} & 14.9672 & 0.5003 & 8.4228 & 16.33 & 2.24 & 79,416 \\
			&  & Zero-DCE++ \cite{Zero-DCE++} & 14.8039 & 0.5161 & 8.3412 & 24.12 & 1.51 & 10,561 \\
			&  & RUAS \cite{RUAS} & 16.4047 & 0.4996 & 5.9297 & - & 8.57 & 3,438 \\
			&  & SCI \cite{SCI} & 14.0226 & 0.5080 & 8.3315 & 2563.68 & {\color{red}1.12} & {\color{red}258} \\
			\cline{2-9}
			& \multirow{3}{*}{Task-irrelevant} & NoiSER-FC (Ours) & {\color{blue}17.5748} & {\color{red}0.7134} & 3.7285 & {\color{blue}1.10} & {\color{blue}1.21} & {\color{blue}1,323} \\
			&  & NoiSER-ES (Ours) & 17.0250 & 0.6563 & {\color{blue}3.7206} & {\color{red}0.58} & {\color{blue}1.21} & {\color{blue}1,323} \\
			&  & NoiSER-Var3 (Ours) & 14.9257 & 0.5998 & {\color{red}3.6806} & {\color{blue}1.10} & {\color{blue}1.21} & {\color{blue}1,323} \\
			\hline
	\end{tabular}}
	\label{tab3}
	\vspace{-3mm}
\end{table*}

\begin{table*}[t]
	\centering
	\caption{Comparison of generalization performance on the LSRW dataset \cite{Dataset-LSRW}, where the best performance is marked in red and the second best one is marked in blue. Clearly, our NoiSER has a stronger generalization ability over different datasets than all other methods.}
	\vspace{-2mm}
	\setlength{\tabcolsep}{2.2mm}{
		\begin{tabular}{c|c|c|ccc|ccc}
			\hline
			& \multirow{2}{*}{Training data} & \multirow{2}{*}{Methods} & \multicolumn{3}{|c|}{LSRW (Huawei)} & \multicolumn{3}{c}{LSRW (Nikon)} \\
			\cline{4-9}
			&  &  & PSNR$\uparrow$ & SSIM$\uparrow$ & NIQE$\downarrow$ & PSNR$\uparrow$ & SSIM$\uparrow$ & NIQE$\downarrow$  \\
			\hline
			\hline
			\multirow{2}{*}{\tabincell{c}{Optimization\\-based}} & \multirow{2}{*}{-} & LIME \cite{LIME} & 15.3376 & 0.4360 & 3.0148 & 14.6362 & 0.3777 & {\color{blue}3.3818} \\
			&  & Zhang et al. \cite{DUAL} & 14.0984 & 0.4327 & {\color{red}2.9228} & 13.0886 & 0.3677 & 3.4620 \\
			\hline
			\multirow{9}{*}{\tabincell{c}{Deep learning\\-based}} & Paired & RetinexNet \cite{RetinexNet} & {\color{blue}16.8127} & 0.3948 & 4.3349 & 13.4853 & 0.2934 & 4.2774 \\
			\cline{2-9}
			& Unpaired & EnlightenGAN \cite{EnlightenGAN} & {\color{red}16.8448} & 0.4832 & 3.0916 & 14.9071 & 0.4065 & {\color{red}3.3745} \\
			\cline{2-9}
			& \multirow{4}{*}{Zero-reference} & Zero-DCE \cite{Zero-DCE} & 14.2002 & 0.3958 & 3.5864 & 11.8197 & 0.3550 & 3.9127 \\
			&  & Zero-DCE++ \cite{Zero-DCE++} & 14.2370 & 0.4163 & 3.4862 & 11.0893 & 0.3675 & 3.7375 \\
			&  & RUAS \cite{RUAS} & 15.6867 & 0.4909 & 3.0399 & 12.1426 & 0.4372 & 3.9902 \\
			&  & SCI \cite{SCI} & 15.2583 & 0.4233 & 3.1263 & 14.4512 & 0.4092 & 3.7864 \\
			\cline{2-9}
			& \multirow{3}{*}{Task-irrelevant} & NoiSER-FC (Ours) & 15.7268 & {\color{red}0.5407} & 3.3555 & {\color{blue}15.5537} & {\color{blue}0.4657} & 3.7784 \\
			&  & NoiSER-ES (Ours) & 15.6968 & {\color{blue}0.5264} & 3.0696 & {\color{red}15.7090} & 0.4584 & 3.4976 \\
			&  & NoiSER-Var3 (Ours) & 15.3001 & 0.5193 & {\color{blue}2.9714} & 15.5260 & {\color{red}0.4672} & 3.5445 \\
			\hline
	\end{tabular}}
	\vspace{-2mm}
	\label{tab4}
\end{table*}

\begin{table}[t]
	\footnotesize
	\centering
	\caption{PSNR comparison between NoiSER and MS-NoiSER. A proper shift can yield a better performance.}
	\vspace{-2mm}
	\setlength{\tabcolsep}{0.4mm}{
		\begin{tabular}{c|ccc|ccc|c}
			\hline
			& R & G & B & LOL & \tabincell{c}{LSRW \\ (Huawei)} & \tabincell{c}{LSRW \\ (Nikon)} & Average \\
			\hline
			\hline
			NoiSER & 0 & 0 & 0 & 17.574 & 15.726 & 15.553 & 16.100 \\
			\hline
			MS-NoiSER & -0.0898 & -0.1412 & -0.1889 & 17.194 & 16.674 & 15.158 & 16.328 \\
			\hline
	\end{tabular}}
	\vspace{-4mm}
	\label{tab5}
\end{table}

\vspace{-1mm}
\section{Experimental Results and Analysis}
\vspace{-0.5mm}
We evaluate the proposed NoiSER for LLIE and illustrate the comparison results with other related methods. 
\label{sec:exper}
\vspace{-1mm}
\subsection{Experimental Descriptions}
\label{subsec:Exp-descript}
\vspace{-0.5mm}
\textbf{Method description.} In general, we use the standard Gaussian distribution $\mathcal{N}(0,1)$ to train NoiSER. Specifically, training NoiSER to full convergence (denoted as NoiSER-FC) can yield better quantitative performance, but the visual effect appears to be obscured by a gray-layer, due to the ability of NoiSER to learn a gray-world mapping. As such, we use an early stopping mechanism (denoted as NoiSER-ES), i.e., stopping training before fully achieving the gray-world mapping, thereby alleviating the gray-layer phenomenon present in NoiSER-FC. However, doing so without fully converging will result in some performance degradation. Besides, from the P-Regression experiment in Fig.\ref{fig:SRDataResult}, it is not hard to see that despite the palette is coarse-granularity, the contrast between adjacent colors is high, and no gray-layer phenomenon occurs. Inspired by the P-Regression, we use Gaussian distribution with higher contrast between adjacent pixels (standard deviation $\sigma=3$) for NoiSER training (denoted NoiSER-Var3). It is found that this can avoid the gray-layer phenomenon and obtain better visual effects compared to both NoiSER-FC and NoiSER-ES, but the performance is somewhat inferior to both NoiSER-FC and NoiSER-ES. Overall, NoiSER-FC possesses the best quantitative performance in reference metrics, NoiSER-Var3 offers the best visual effects, and NoiSER-ES provides a good trade-off between visual effect and quantitative performance, which allows for flexible selection based on specific needs in real scenarios. In the following Subsections, we conduct extensive experiments to evaluate these three versions of NoiSER.

\begin{figure*}[t]
	\centering
	\includegraphics[width=2\columnwidth]{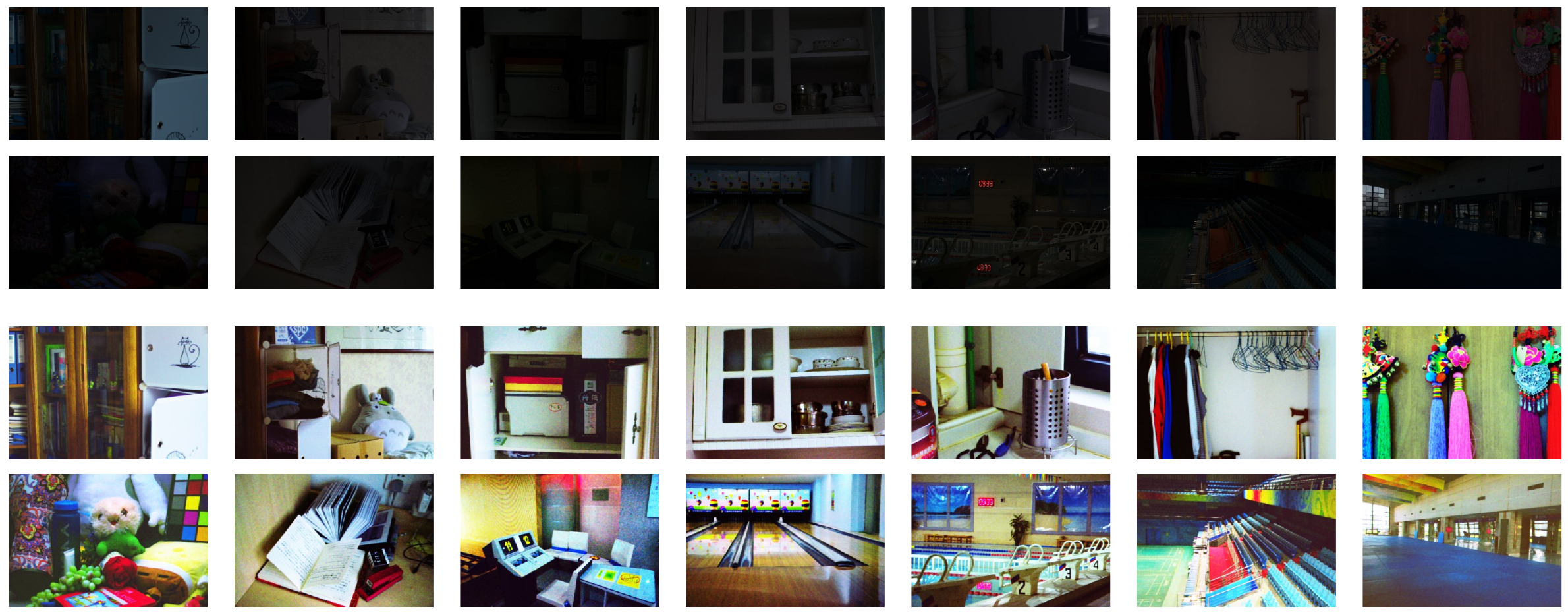}
	\vspace{-2mm}
	\caption{Intuitive visual display of our NoiSER-Var3 on the widely-used LOL dataset \cite{RetinexNet}. We exclude ``23.png" due to its similarity to ``22.png", facilitating a two-line display. Just a simple and straightforward noise self-regression can yield a surprising visual effect.}
	\label{fig:DisplayLOL}
	\vspace{-2mm}
\end{figure*}

\begin{figure*}[t]
	\centering
	\includegraphics[width=2\columnwidth]{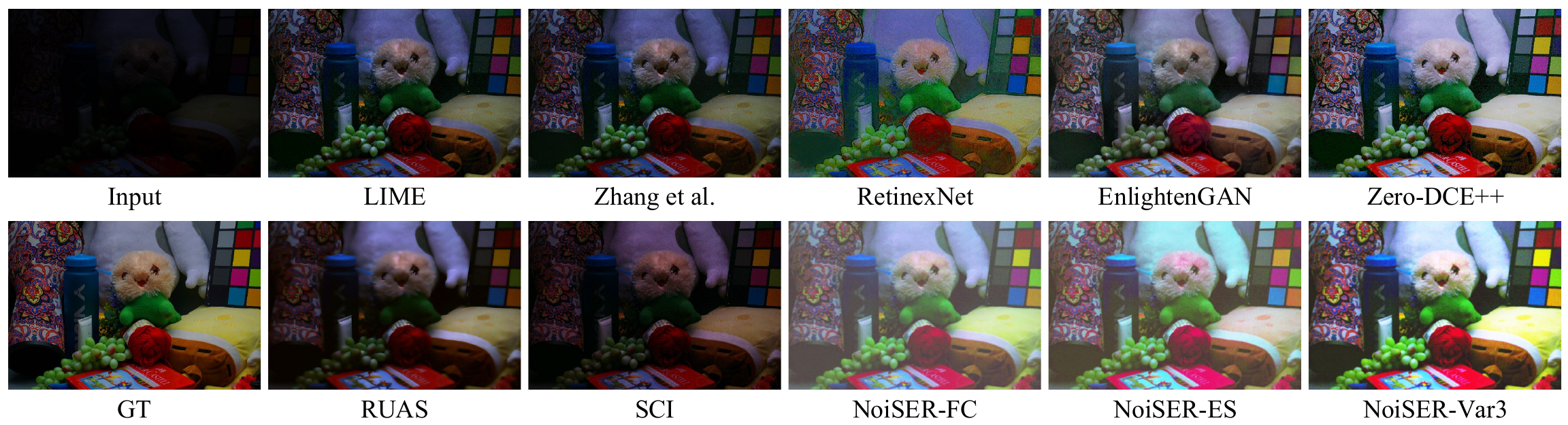}
	\vspace{-2mm}
	\caption{Visual results of different image enhancement methods based on the LOL dataset \cite{RetinexNet}, including LIME \cite{LIME}, Zhang et al. \cite{DUAL}, RetinexNet \cite{RetinexNet}, EnlightenGAN \cite{EnlightenGAN}, Zero-DCE++ \cite{Zero-DCE++}, RUAS \cite{RUAS}, SCI \cite{SCI} and our NoiSER. Clearly, our NoiSER obtains better visualization effects, despite no task-related data have been used for training. Specifically, our NoiSER can enhance the image contents more visibly and naturally, even beyond the ground-truth, which is attributed to the ability of our method to learn a gray-world mapping.}
	\label{fig:VisualLOL}
	\vspace{-2mm}
\end{figure*}

\textbf{Comparison description.} Since our NoiSER does not use any task-related data for training, all the tests on various datasets are equivalent to evaluating the generalization ability, instead of fitting ability. Training and testing on the same dataset is apparently against our intention, however we still use the most widely-used LOL dataset \cite{RetinexNet} for training \& testing to show that our method is competitive.

\vspace{-1mm}
\subsection{Experimental Settings}
\label{subsec:Exp-settings}
\textbf{Evaluated datasets.} We conduct experiments on two widely-used low-light image datasets (i.e., LOL \cite{RetinexNet} and LSRW \cite{Dataset-LSRW}) and a multi-exposure dataset (i.e., SICE \cite{Dataset-SICE}). We train and test other competitors on LOL dataset (note that our NoiSER always uses noise), and directly test them on LSRW using the pretrained model on LOL. Besides, we use SICE to measure the over-exposure suppression capability. The details of the three datasets are as follows:
\begin{itemize}
	\vspace{-1mm}
	\item \textbf{LOL}: LOL is the most widely-used dataset for LLIE task, including 485 training pairs and 15 testing pairs with resolution 400$\times$600.
	\item \textbf{LSRW}: LSRW has two subsets: LSRW (Huawei) and LSRW (Nikon). LSRW (Huawei) has 2,480 training pairs and 30 testing pairs with resolution 720$\times$960, while LSRW (Nikon) contains 3,170 training pairs and 20 testing pairs with resolution 640$\times$960.
	\item \textbf{SICE}: SICE includes 4,413 multi-exposure images with resolution between 3000$\times$2000 and 6000$\times$4000. It is divided into two parts, namely, ``Dataset\_Part1" and ``Dataset\_Part2", and we only adopt the overexposed image named ``7.JPG" from ``Dataset\_Part2" as an example in our experiments.	
\end{itemize}

\textbf{Evaluation metrics.} We use three widely-used performance metrics to evaluate the quantitative result of each method, i.e., PSNR, SSIM \cite{Metric-SSIM}) and NIQE \cite{Metric-NIQE}. Besides, we also use three application metrics to evaluate the possibility of each method for practical deployment and application, i.e., training time (TT, in minutes), inference time (IT, in milliseconds) and number of parameters (No.P). In all experiments, we use $\uparrow$ to indicate the higher the better, and use $\downarrow$ to indicate the lower the better.

\textbf{Compared methods.} Two optimization-based methods (LIME \cite{LIME} and Zhang et al. \cite{DUAL}) and six deep learning-based methods are included for comparison. For deep methods, different types of training data are leveraged to fit the model, i.e., paired (RetinexNet \cite{RetinexNet}), unpaired data (EnlightenGAN \cite{EnlightenGAN}) and zero-reference data (Zero-DCE \cite{Zero-DCE}, Zero-DCE++ \cite{Zero-DCE++}, RUAS \cite{RUAS} and SCI \cite{SCI}). Note that we mainly compare with the four zero-reference methods, as they are closer to our NoiSER in terms of data constraints.

\begin{figure*}[t]
	\centering
	\includegraphics[width=2\columnwidth]{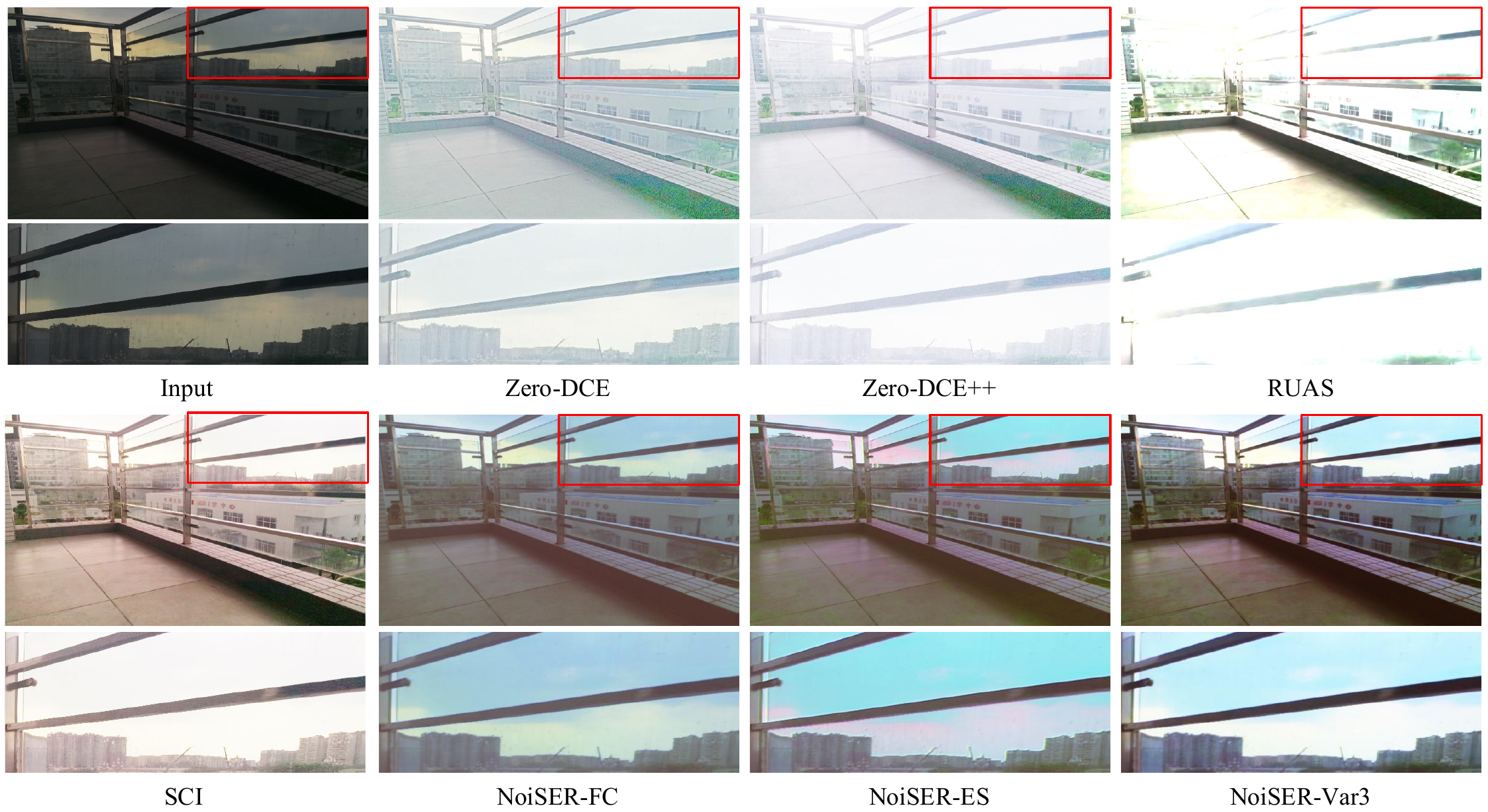}
	\vspace{-2mm}
	\caption{Enhancement results of each method on the LSRW (Huawei) dataset \cite{Dataset-LSRW}, including Zero-DCE \cite{Zero-DCE}, Zero-DCE++ \cite{Zero-DCE++}, RUAS \cite{RUAS}, SCI \cite{SCI} and our NoiSER. Clearly, all other compared methods overexpose the brighter areas of the images, while our NoiSER effectively suppresses the exposure and does a good job in recovering the texture detail of the image.}
	\label{fig:VisualLSRWHuawei}
	\vspace{-3mm}
\end{figure*}

\begin{figure*}[t]
	\centering
	\includegraphics[width=2\columnwidth]{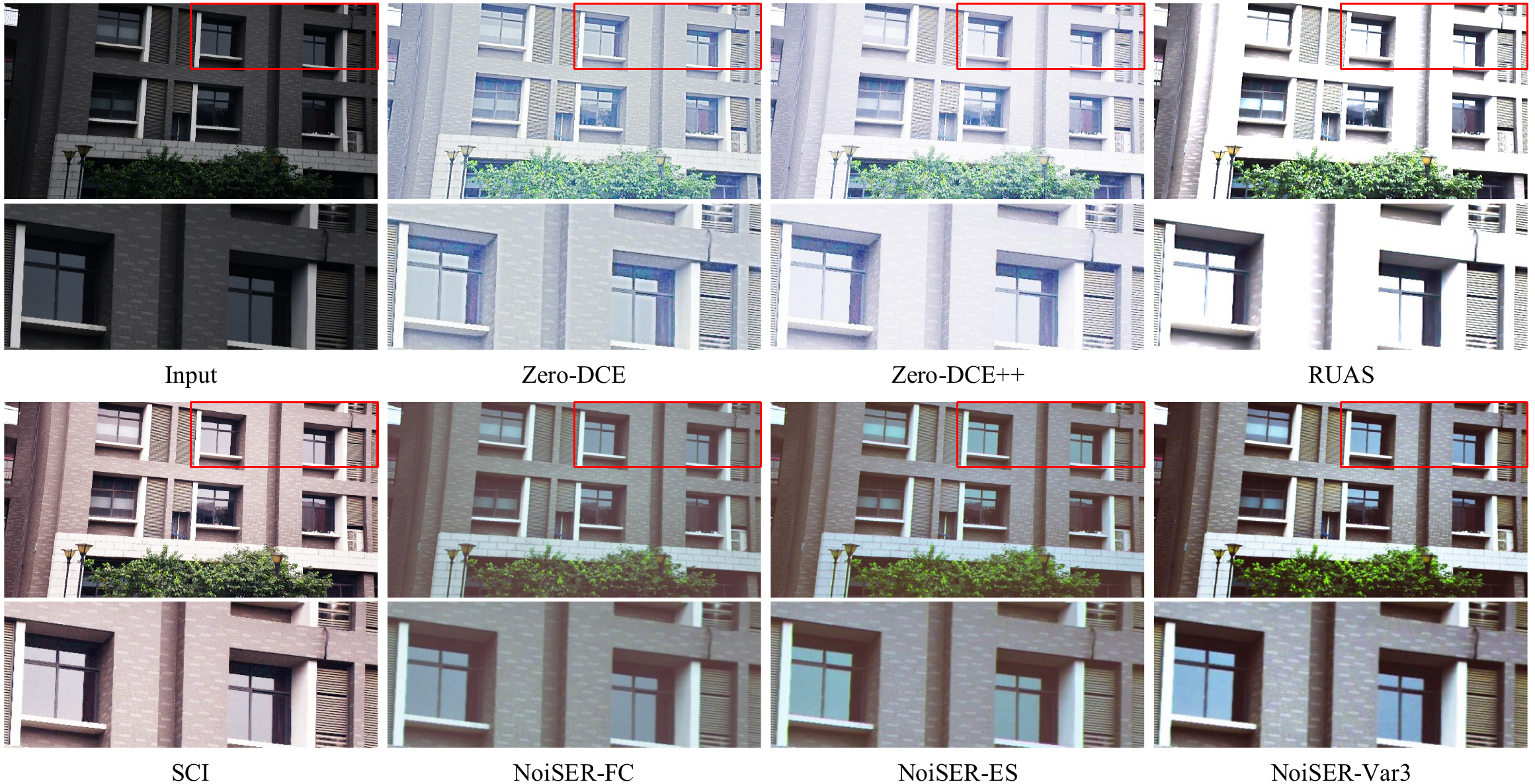}
	\vspace{-2mm}
	\caption{Visual results of different enhancement methods based on the LSRW (Nikon) dataset \cite{Dataset-LSRW}, including Zero-DCE \cite{Zero-DCE}, Zero-DCE++ \cite{Zero-DCE++}, RUAS \cite{RUAS}, SCI \cite{SCI} and our NoiSER. Because the image itself is not very dark, all other compared methods either overexpose the image or distort the color, while our NoiSER achieves a better visual result.}
	\label{fig:VisualLSRWNikon}
	\vspace{-4mm}
\end{figure*}

\textbf{Implementation details.} Based on PyTorch 1.10.1 and Python 3.6.9, we train and evaluate our NoiSER on single NVIDIA RTX 2080 Ti GPU. We train our NoiSER for 2000 iterations (600 for NoiSER-ES) with a batch size of 1 and a fixed learning rate of 2e-4. We sample the noise from standard Gaussian distribution $\mathcal{N}(0,1)$ ($\mathcal{N}(0,3)$ for NoiSER-Var3) with a shape of $104\times104$ for training. Besides, the Adam optimizer \cite{Adam} is utilized for training with $\beta_1=0.5$ and $\beta_2=0.999$. To alleviate the noise in the enhanced results, we add TV regularization during training. The total loss function of NoiSER is defined as 
\begin{equation}\label{eqn11}
	\mathcal{L}_{total} = \mathcal{L}_{l_1} + \mathcal{L}_{tv}. 
\end{equation}

\vspace{-1mm}
\subsection{Quantitative Evaluations}
\subsubsection{Results on LOL Dataset}
We first examine the fitting ability of each method on the LOL dataset in Table \ref{tab3}. For the TT metric, we get the results on a single NVIDIA RTX 2080 Ti using the default configuration according to the codes provided by authors. We do not list the training time of RUAS \cite{RUAS}, since it involves a process of neural architecture search. We find that: 1) optimization-based methods take longer inference time for the use of CPU for computation, which is a major impediment to real applications; 2) when training with paired/unpaired data, RetinexNet \cite{RetinexNet} and EnlightenGAN \cite{EnlightenGAN} yield better results due to the relatively strong constraints, but they have a larger number of parameters and slower inference speed; 3) the zero-reference methods require minimal computational resources, which are relatively efficient and have a stronger application potential; 4) although NoiSER does not use any task-related data for training, it is still highly competitive to the other methods across all metrics. From the comparison to the zero-reference methods, it can be seen that our NoiSER perfectly outperforms them all with significantly less training time. 

\vspace{-1mm}
\subsubsection{Results on LSRW Dataset}
\vspace{-0.5mm}
To test the generalization ability, we evaluate the methods on two test subsets,  LSRW (Huawei) and LSRW (Nikon). The numerical results are shown in Table \ref{tab4}. We can find that: 1) the models fitted using the task-related data fail to generalize well to all datasets, since they perform well on some datasets but poorly on others. For example, EnlightenGAN \cite{EnlightenGAN} and RetinexNet \cite{RetinexNet} obtain the best records on LSRW (Huawei), but worse on LSRW (Nikon), which is inevitable for the usage of task-related data; 2) our NoiSER generalizes generally well and is more balanced\footnote{It means that NoiSER generalizes stably on different datasets, i.e., it is less likely to be extremely good on one and extremely poor on the other.} across different datasets, indicating our approach is closer to the essence of low-light enhancement; 3) from the comparison to those zero-reference methods that are more close to ours, our NoiSER obtains significant superiority in all metrics. 

\vspace{-1mm}
\subsubsection{Results for MS-NoiSER Optimization}
Instead of using zero-mean Gaussian noise, we adopt mean-shifted Gaussian noise for self-regression training to prove the effectiveness of the proposed MS-NoiSER strategy. The quantitative results are shown in Table \ref{tab5}, from which we see that: 1) a proper offset may yield better performance; 2) compared to NoiSER, MS-NoiSER can help those datasets improve the performance whose RGB channel mean is far from central-grey. We can also obtain a inspiration, i.e., MS-NoiSER provides a way to arbitrarily adjust the brightness of the LLIE result using different Gaussian mean.
\begin{figure*}[t]
	\centering
	\includegraphics[width=2\columnwidth]{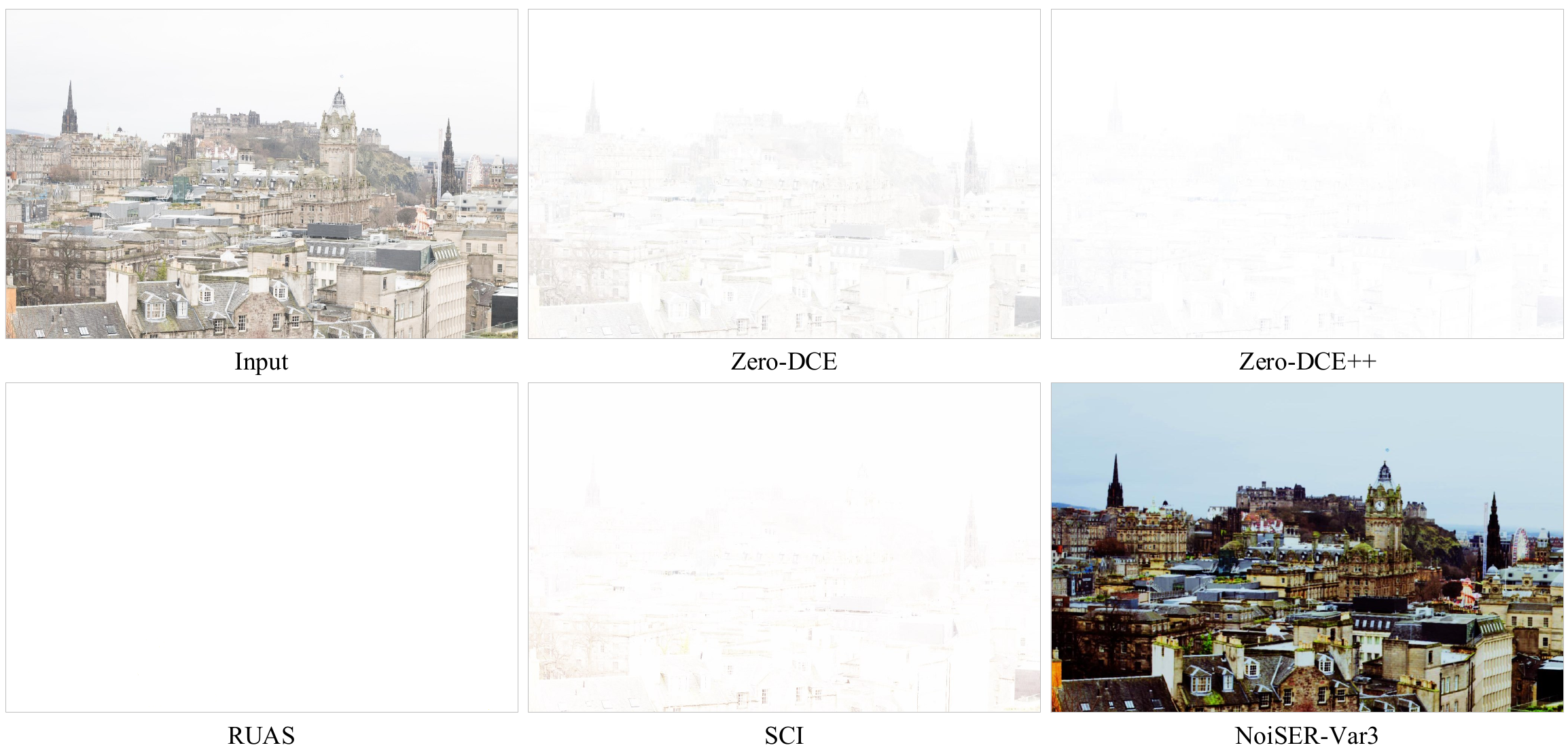}
	\caption{Comparison of the overexposure suppression effects of each method on the SICE dataset \cite{Dataset-SICE}, including Zero-DCE \cite{Zero-DCE}, Zero-DCE++ \cite{Zero-DCE++}, RUAS \cite{RUAS}, SCI \cite{SCI} and our NoiSER. We see that all other competitors failed to handle the overexposed inputs appropriately, while our NoiSER has an inborn automated exposure suppression capability.}
	\label{fig:OverexposureSuppression}
\end{figure*}

\vspace{-1mm}
\subsection{Visualization Evaluations}
\subsubsection{Visual Analysis on LOL Dataset}
In addition to the numerical results, we also display some visualization results for observation. NoiSER is easy to implement and can yield surprising results. Considering that the test set of LOL only has 15 low-light images, we show the entire set to give a more intuitive view for the enhancement effect in Fig.\ref{fig:DisplayLOL}. Clearly, by simply performing a noise self-regression for training, all low-light images can be enhanced with rich texture, content and color information.

We then compare each method on an extremely dark image (from which human eyes hardly see anything) of LOL \cite{RetinexNet} dataset in Fig.\ref{fig:VisualLOL}. We see that: 1) both optimization-based and deep LLIE methods using task-related data can enhance this very dark image to some extent, however the restored images are still inferior in terms of illumination enhancement, detail recovery and color preservation; 2) our NoiSER enhances this overdark image considerably, and the restored image is even beyond the brightness and content naturalness of the ground-truth, which can be attributed to the learned gray-world mapping that always forces the channel means of low-light image to be close to the central grey; 3) for the visual effects of several variants of our NoiSER, NoiSER-FC $<$ NoiSER-ES $<$ NoiSER-Var3, which is consistent with Section \ref{subsec:Exp-descript}, i.e., NoiSER-FC seems to be covered by a grey layer, which should be caused by the learned gray-world mapping, and NoiSER-ES shows that early stopping mechanism can alleviate this issue. To obtain the most visually pleasing effect, we can adopt the manner of increasing the variance of Gaussian distribution, since NoiSER-Var3 clearly shows a better visual effect.

\vspace{-1mm}
\subsubsection{Visual Analysis on LSRW Dataset}
To evaluate the generalization ability by visual analysis, we compare the visual results of each method on LSRW \cite{Dataset-LSRW} dataset in Fig.\ref{fig:VisualLSRWHuawei} and Fig.\ref{fig:VisualLSRWNikon}. We can see that: 1) when the input image is not very dark, all the zero-reference methods tend to produce overexposure enhanced images, which highlights the stringency of the task-related methods to training data; 2) owing to the simple and straightforward training process, our NoiSER has a powerful generalization ability. From the comparison of the local details in Fig.\ref{fig:VisualLSRWHuawei} and Fig.\ref{fig:VisualLSRWNikon}, it is clear that our NoiSER can recover the texture of the images more accurately and get better visual effects.

\begin{figure*}[t]
	\centering
	\includegraphics[width=2\columnwidth]{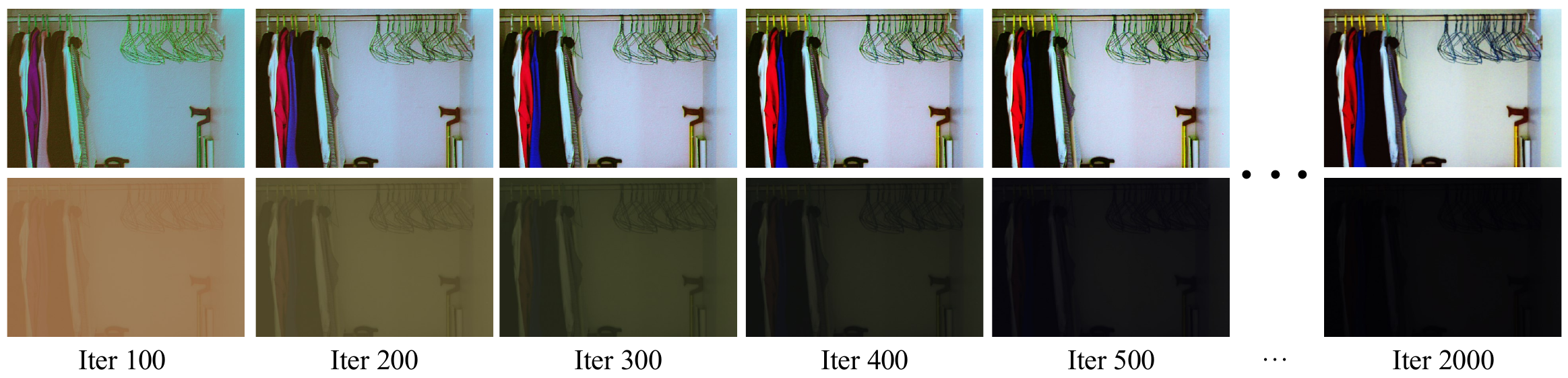}
	\vspace{-2mm}
	\caption{Visual ablation study on the instance normalization in SRM. Top row denotes the training process of NoiSER-Var3, while the bottom row denotes the training process of NoiSER-Var3 without IN. Clearly, the enhancement effect becomes very poor when IN is removed, i.e., IN is very important to naturally remediate the overall magnitude/lighting of the input image in our model.}
	\label{fig:Ablation_IN}
	\vspace{-4mm}
\end{figure*}

\begin{figure*}[t]
	\centering
	\includegraphics[width=2\columnwidth]{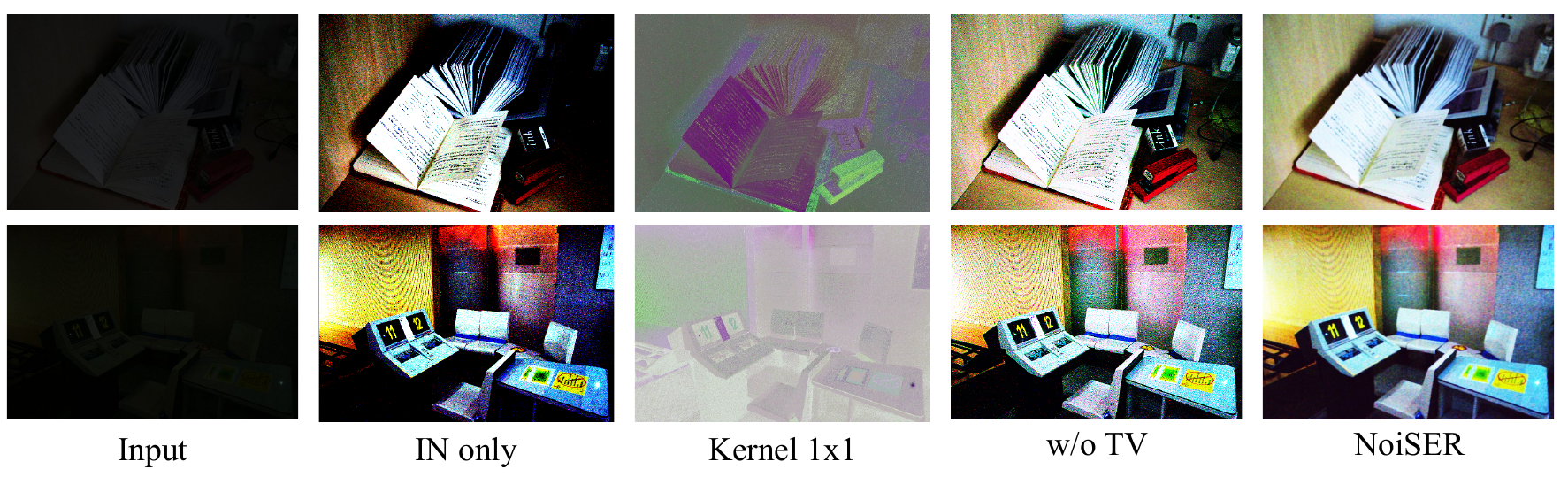}
	\vspace{-2mm}
	\caption{Visual ablation studies of 1) infeasibility of solely using IN, 2) Effect of kernel size and 3) effectiveness of TV.}
	\label{fig:Ablations}
	\vspace{-2mm}
\end{figure*}

\begin{figure*}[t]
	\centering
	\includegraphics[width=2\columnwidth]{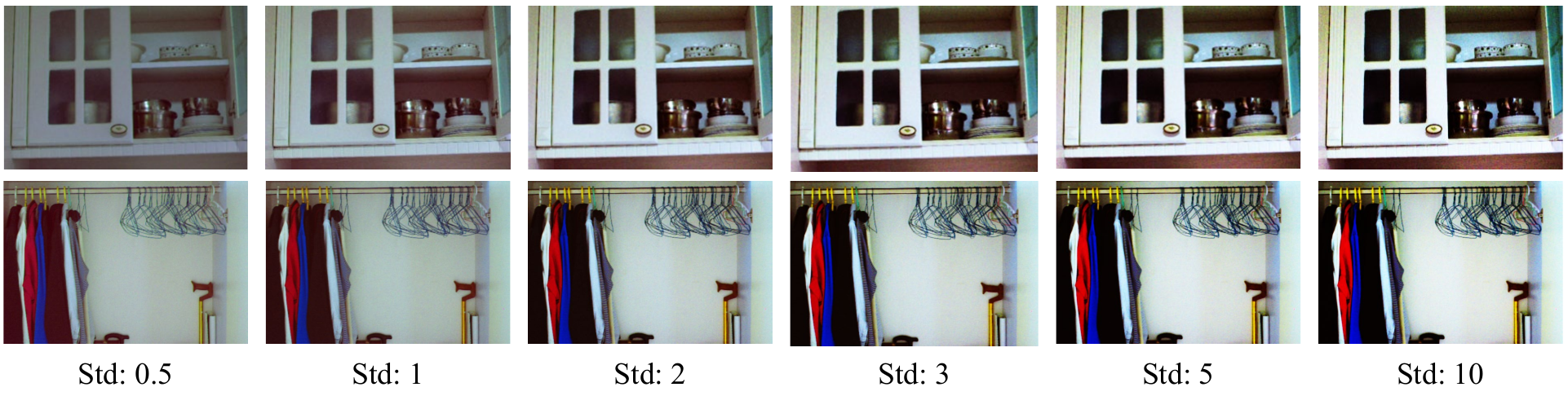}
	\vspace{-2mm}
	\caption{Visual ablation study on different standard deviations. As can be seen, increasing the standard deviation can make the image brighter. Therefore, the brightness can be customized by adjusting the standard deviation.}
	\label{fig:Ablation_std}
	\vspace{-2mm}
\end{figure*}

\vspace{-1mm}
\subsection{Automated Overexposure Suppression}
Finally, we evaluate each method to process the overexposed image of the SICE dataset. As shown in Fig.\ref{fig:OverexposureSuppression}, existing methods Zero-DCE \cite{Zero-DCE}, Zero-DCE++ \cite{Zero-DCE++}, RUAS \cite{RUAS} and SCI \cite{SCI} obtain poor results and the processed images are almost completely corrupted, since their aim is to learn a mapping from low to high illumination. One possible solution is to use a large scale multi-exposure dataset to train the model, so that the model can handle both overdark and overexposed images. However, it is still tricky to handle the extremely complex illumination in reality. Therefore, we are delighted to say that our NoiSER has an inborn ability to suppress the overexposure, since it learns a gray-world mapping which can automatically turn extreme light or dark into moderates, as shown in ``NoiSER-Var3" in Fig.\ref{fig:OverexposureSuppression}.

\begin{figure}[t]
	\centering
	\includegraphics[width=1\columnwidth]{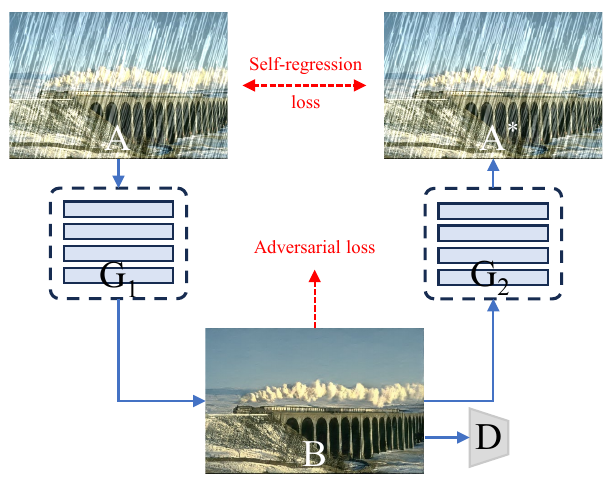}
	\vspace{-2mm}
	\caption{Unsupervised deraining architecture. $G_1$ and $G_2$ are generators equipped with IN, and the output range of generator is limited to [-1,1]. Although the entire architecture does not have additional low-light enhancement data and losses, after training, the pretrained model $G_1$ can still enhance low-light images.}
	\label{fig:Deraining_arch}
	\vspace{-4mm}
\end{figure}

\begin{figure}[t]
	\centering
	\includegraphics[width=1\columnwidth]{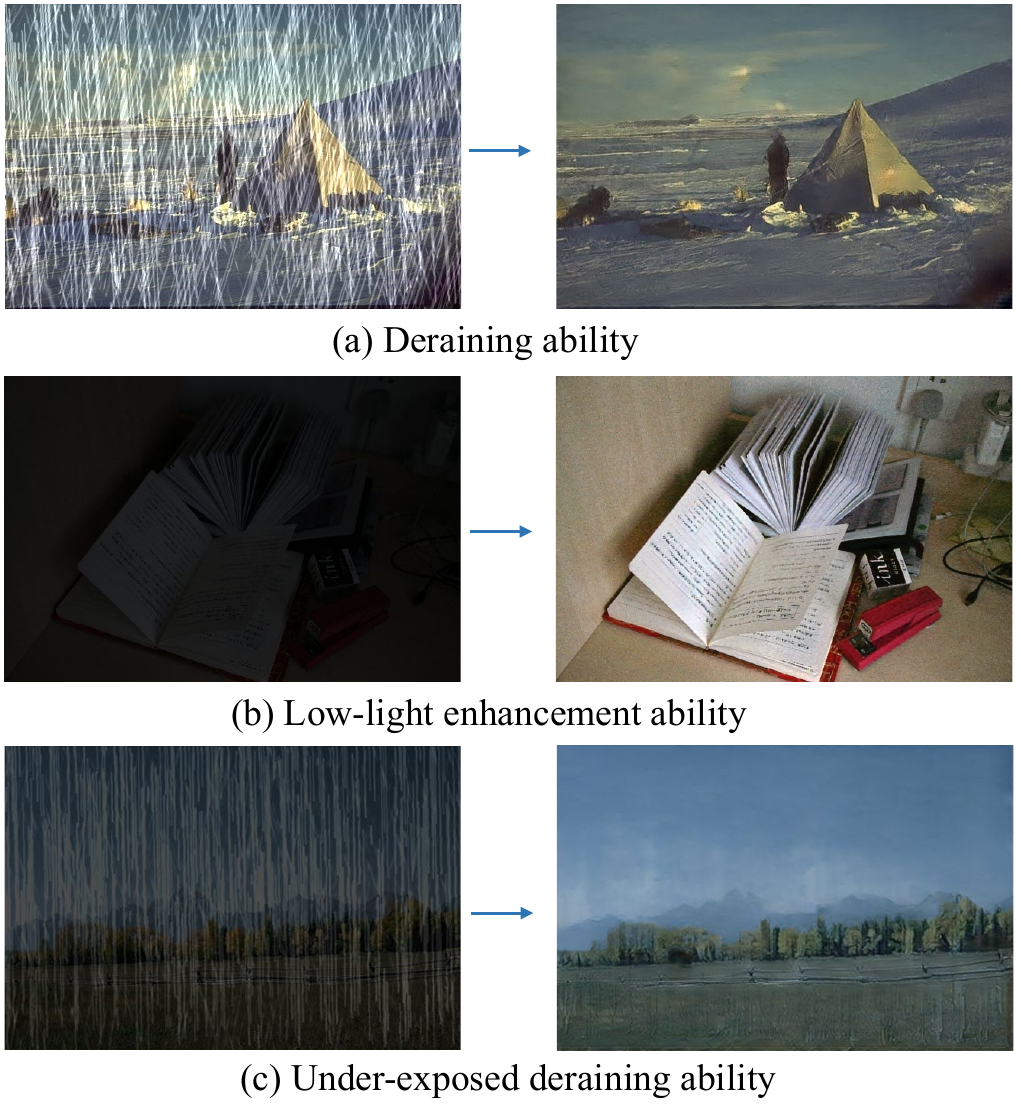}
	\vspace{-2mm}
	\caption{After training on deraining dataset Rain100H \cite{Rain100H}, the pretrained model has the ability to process: 1) deraining, 2) low-light enhancement and 3) under-exposed deraining.}
	\label{fig:Deraining_results}
	\vspace{-6mm}
\end{figure}

\vspace{-1mm}
\subsection{Ablation Studies}
\textbf{1) Effectiveness of instance normalization.} To evaluate the importance and the effect of the instance-normalization layer in remediating the overall magnitude/lighting of the input image, we conduct experiments by removing the IN with other settings unchanged, and the visual effects are displayed in Fig.\ref{fig:Ablation_IN}. As can be seen, after removing the IN, NoiSER cannot work normally, which indicates the importance of the IN layer in our model.


\noindent\textbf{2) Effectiveness of TV regularization.} To verify the assistance of TV regularization to noise alleviation, we conduct experiments by removing the TV regularization with other settings unchanged, and the visual effects are displayed in Fig.\ref{fig:Ablations}. As can be seen from the Figure, after removing the TV regularization, the visualization results of NoiSER-Var3 suffer from severe noise, which indicates the importance of the TV regularization.

\noindent\textbf{3) Infeasibility of solely using IN.} To confirm that NoiSER's capability to enhance an image isn't solely due to the IN layer, we try to process low-light images using only IN. The visual results are displayed in Fig.\ref{fig:Ablations}. These results demonstrate that using only the IN without the whole NoiSER does not enhance low-light images well, highlighting the necessity of our NoiSER technique. 

\noindent\textbf{4) Effect of kernel size.} To evaluate the effect of different kernel sizes in CNN on the performance, we conducted relevant experiments using 1$\times$1 convolution to replace the original 3$\times$3 convolution, and the experimental results are shown in Fig.\ref{fig:Ablations}. From the figure, it can be seen that the parameter space of the 1x1 convolution is too small to allow CNN to learn a good enhancement ability.

\noindent\textbf{5) Effect of standard deviations.} To evaluate the effect of different standard deviations of random noise on the performance, we conducted six groups of experiments, i.e., setting standard deviations to 0.5, 1, 2, 3, 5, 10, respectively, and the experimental results are shown in Fig.\ref{fig:Ablation_std}. From the figure, it can be seen that a higher standard deviation makes the enhanced image brighter, which means that if unexpected overexposure occurs, it can be mitigated by lowering the standard deviation.

\subsection{Extension Experiments}

To verify the potential ability of NoiSER on joint tasks, we chose the deraining task to explore how to integrate NoiSER technique into the unsupervised deraining in an elegant way. To this end, we designed an architecture as shown in Fig.\ref{fig:Deraining_arch}, and the whole architecture includes three key components, i.e., two generators ($G_1$ and $G_2$) and one discriminator $D$. The architecture tries to achieve a cycle, i.e., $A\rightarrow B\rightarrow A^*\approx A$, where $A$ denotes rainy images, $B$ denotes restored rain-free images and $A^*$ denotes reconstructed rainy images, in which the NoiSER technique is implicitly integrated. Specifically, considering the three essential conditions in NoiSER: 1) noise self-regression process, 2) IN-equipped CNN, 3) output mapping interval of [-1,1]. The latter two conditions can be easily satisfied since the generators $G_1$ and $G_2$ have been equipped with IN and the output range of generators has been limited to [-1,1]. For condition 1), it is difficult to directly integrate it into deraining task, so we take an approximate approach, as shown in Fig.\ref{fig:Deraining_arch}, we directly regard the rainy images as the ``noise" in noise self-regression and input it to the generator $G_1$ instead of adding a new noise input. Besides, we treat the entire architectural cycle $A\rightarrow B\rightarrow A^*\approx A$ as an indirect self-regression process rather than adding a direct $A\rightarrow A^*\approx A$ self-regression process.

Based on the designed architecture, we conduct relevant experiments on Rain100H \cite{Rain100H} datasets. It is worth noting that only unpaired deraining data (regarded as ``noise" in NoiSER) is used in the entire training process, resulting pretrained model has the ability to directly enhance low-light images. Besides, we further verified whether the pretrained model directly has the ability to process under-exposed rainy data. Firstly, due to the lack of under-exposed rainy data, we adjusted the global brightness based on Rain100H to synthesize under-exposed rainy data. Then we directly performed inference on this dataset, and the results showed that the trained model also has the ability to handle under-exposed rainy data. The visualization results are shown in Fig.\ref{fig:Deraining_results}, which demonstrates that the designed architecture has the abilities of vanilla NoiSER, and also proves the potential ability of NoiSER in handling joint tasks.

\subsection{Discussion and Limitations}

NoiSER has a significant difference from other data-driven deep LLIE methods (paired, unpaired and zero-reference), outperforming the zero-reference methods that are closest to our approach at the data level. On the whole, the advantages of NoiSER can be summarized as: lightweight, efficient, strong generalization ability, automatic exposure suppression, and the ability to handle joint tasks with other tasks. However, everything has two sides, and NoiSER also has its limitations. We summarize the limitations in two points as follows:

1) The limitation for scene: Despite our method having stronger generalization capabilities, it is still difficult to handle some special cases, such as dimly lit bars filled with colorful lights. In fact, this is also a relatively complex and hard real-world scenario, and other existing types of methods also cannot well handle this scenario due to the weak generalization capabilities.

2) The limitation for task: Deep learning models usually have task-generality, i.e., a same model structure can be trained on different datasets to complete different tasks. In contrast, our NoiSER is tightly coupled with the task of low-light image enhancement and cannot independently complete other tasks. However, NoiSER can be combined with other tasks for joint processing, such as joint deraining and LLIE, which alleviates this limitation to some extent.

\vspace{-1mm}
\section{Conclusion}
\label{sec:concl}
We have discussed a new problem, i.e., how to enhance a low-light image by deep learning without any task-related data. We prove that the problem can easily solved by a straightforward noise self-regression approach that learns a simple convolutional neural network equipped with a instance-normalization layer by taking random noise as the input for training. Extensive experiments show our method can obtain highly-competitive and even better performance than current SOTA methods with different task-related data, in terms of enhancement ability, stable generalization capability, automated exposure suppression and negligible computational consumption.
In the future, we will further think about how to optimize the noise for improving visual effect. Besides, considering the characteristics of NoiSER which does not require any task-related data, we will further explore the possible combination for NoiSER to other low-level vision tasks, which will elegantly enable joint processing of LLIE and other low-level vision tasks.

\section*{Acknowledgment}
This work is supported by the National Natural Science Foundation of China (62472137, 62072151), Anhui Provincial Natural Science Fund for the Distinguished Young Scholars (2008085J30), Open Foundation of Yunnan Key Laboratory of Software Engineering (2023SE103),  CCF-Baidu Open Fund (CCF-BAIDU202321) and CAAI-Huawei MindSpore Open Fund (CAAIXSJLJJ-2022-057A). Corresponding authors: Suiyi Zhao and Meng Wang.

{\small
	\bibliographystyle{ieee_fullname}
	\bibliography{NoiSER}

\begin{thebibliography}{10}\itemsep=-1pt

\bibitem{InductionReasoning1}
Pierpaolo Battigalli and Marciano~M. Siniscalchi.
\newblock Strong belief and forward induction reasoning.
\newblock {\em J. Econ. Theory}, 106(2):356--391, 2002.

\bibitem{Gary-world_Hypothesis}
Gershon Buchsbaum.
\newblock A spatial processor model for object colour perception.
\newblock {\em Journal of the Franklin institute}, 310(1), 1980.

\bibitem{Dataset-SICE}
Jianrui Cai, Shuhang Gu, and Lei Zhang.
\newblock Learning a deep single image contrast enhancer from multi-exposure
  images.
\newblock {\em {IEEE} Trans. Image Process.}, 27:2049--2062, 2018.

\bibitem{Retinexformer}
Yuanhao Cai, Hao Bian, Jing Lin, Haoqian Wang, Radu Timofte, and Yulun Zhang.
\newblock Retinexformer: One-stage retinex-based transformer for low-light
  image enhancement.
\newblock In {\em Proceedings of the {IEEE/CVF} International Conference on
  Computer Vision, Paris, France}, pages 12470--12479, 2023.

\bibitem{HE1}
Turgay {\c{C}}elik and Tardi Tjahjadi.
\newblock Contextual and variational contrast enhancement.
\newblock {\em {IEEE} Trans. Image Process.}, 20(12), 2011.

\bibitem{Low_Light_Object_Detection}
Ziteng Cui, Guo{-}Jun Qi, Lin Gu, Shaodi You, Zenghui Zhang, and Tatsuya
  Harada.
\newblock Multitask {AET} with orthogonal tangent regularity for dark object
  detection.
\newblock In {\em Proceedings of the {IEEE/CVF} International Conference on
  Computer Vision, Montreal, QC, Canada}, pages 2533--2542, 2021.

\bibitem{Low_Light_Semantic_Segmentation}
Dengxin Dai and Luc~Van Gool.
\newblock Dark model adaptation: Semantic image segmentation from daytime to
  nighttime.
\newblock In {\em Proceedings of the {IEEE} International Conference on
  Intelligent Transportation Systems, Maui, HI, USA, November 4-7, 2018}, pages
  3819--3824, 2018.

\bibitem{PairLIE}
Zhenqi Fu, Yan Yang, Xiaotong Tu, Yue Huang, Xinghao Ding, and Kai{-}Kuang Ma.
\newblock Learning a simple low-light image enhancer from paired low-light
  instances.
\newblock In {\em Proceedings of the {IEEE/CVF} Conference on Computer Vision
  and Pattern Recognition, Vancouver, BC, Canada}, pages 22252--22261, 2023.

\bibitem{GAN}
Ian~J. Goodfellow, Jean Pouget{-}Abadie, Mehdi Mirza, Bing Xu, David
  Warde{-}Farley, Sherjil Ozair, Aaron~C. Courville, and Yoshua Bengio.
\newblock Generative adversarial networks.
\newblock {\em Commun. {ACM}}, 63(11):139--144, 2020.

\bibitem{Zero-DCE}
Chunle Guo, Chongyi Li, Jichang Guo, Chen~Change Loy, Junhui Hou, Sam Kwong,
  and Runmin Cong.
\newblock Zero-reference deep curve estimation for low-light image enhancement.
\newblock In {\em Proceedings of the Conference on Computer Vision and Pattern
  Recognition, Seattle, WA, USA}, pages 1777--1786, 2020.

\bibitem{LIME}
Xiaojie Guo.
\newblock {LIME:} {A} method for low-light image enhancement.
\newblock In {\em Proceedings of the {ACM} Conference on Multimedia Conference,
  Amsterdam, The Netherlands}, pages 87--91, 2016.

\bibitem{Dataset-LSRW}
Jiang Hai, Zhu Xuan, Ren Yang, Yutong Hao, Fengzhu Zou, Fang Lin, and Songchen
  Han.
\newblock R2rnet: Low-light image enhancement via real-low to real-normal
  network.
\newblock {\em J. Vis. Commun. Image Represent.}, 90:103712, 2023.

\bibitem{InductionReasoning2}
Evan Heit and Caren~M Rotello.
\newblock Relations between inductive reasoning and deductive reasoning.
\newblock {\em Journal of Experimental Psychology: Learning, Memory, and
  Cognition}, 36(3):805, 2010.

\bibitem{Auto-Encoder}
Geoffrey~E Hinton and Ruslan~R Salakhutdinov.
\newblock Reducing the dimensionality of data with neural networks.
\newblock {\em science}, 313(5786):504--507, 2006.

\bibitem{EnlightenGAN}
Yifan Jiang, Xinyu Gong, Ding Liu, Yu Cheng, Chen Fang, Xiaohui Shen, Jianchao
  Yang, Pan Zhou, and Zhangyang Wang.
\newblock Enlightengan: Deep light enhancement without paired supervision.
\newblock {\em {IEEE} Trans. Image Process.}, 30:2340--2349, 2021.

\bibitem{Retinex-Based1}
Daniel~J. Jobson, Zia{-}ur Rahman, and Glenn~A. Woodell.
\newblock A multiscale retinex for bridging the gap between color images and
  the human observation of scenes.
\newblock {\em {IEEE} Trans. Image Process.}, 6(7), 1997.

\bibitem{Adam}
Diederik~P. Kingma and Jimmy Ba.
\newblock Adam: {A} method for stochastic optimization.
\newblock In {\em Proceedings of the International Conference on Learning
  Representations, San Diego, CA, USA}, 2015.

\bibitem{VAE}
Diederik~P. Kingma and Max Welling.
\newblock Auto-encoding variational bayes.
\newblock In {\em Proceedings of the International Conference on Learning
  Representations, Banff, AB, Canada}, 2014.

\bibitem{HE2}
Chulwoo Lee, Chul Lee, and Chang{-}Su Kim.
\newblock Contrast enhancement based on layered difference representation of 2d
  histograms.
\newblock {\em {IEEE} Trans. Image Process.}, 22(12):5372--5384, 2013.

\bibitem{Zero-DCE++}
Chongyi Li, Chunle Guo, and Chen~Change Loy.
\newblock Learning to enhance low-light image via zero-reference deep curve
  estimation.
\newblock {\em {IEEE} Trans. Pattern Anal. Mach. Intell.}, 44(8):4225--4238,
  2022.

\bibitem{SCL_LLE}
Dong Liang, Ling Li, Mingqiang Wei, Shuo Yang, Liyan Zhang, Wenhan Yang, Yun
  Du, and Huiyu Zhou.
\newblock Semantically contrastive learning for low-light image enhancement.
\newblock In {\em Proceedings of the {AAAI} Conference on Artificial
  Intelligence, Virtual}, pages 1555--1563, 2022.

\bibitem{Low_Light_Face_Detection_2}
Jinxiu Liang, Jingwen Wang, Yuhui Quan, Tianyi Chen, Jiaying Liu, Haibin Ling,
  and Yong Xu.
\newblock Recurrent exposure generation for low-light face detection.
\newblock {\em {IEEE} Trans. Multim.}, 24:1609--1621, 2022.

\bibitem{RUAS}
Risheng Liu, Long Ma, Jiaao Zhang, Xin Fan, and Zhongxuan Luo.
\newblock Retinex-inspired unrolling with cooperative prior architecture search
  for low-light image enhancement.
\newblock In {\em Proceedings of the {IEEE} Conference on Computer Vision and
  Pattern Recognition, virtual}, pages 10561--10570, 2021.

\bibitem{LLNet}
Kin~Gwn Lore, Adedotun Akintayo, and Soumik Sarkar.
\newblock Llnet: {A} deep autoencoder approach to natural low-light image
  enhancement.
\newblock {\em Pattern Recognit.}, 61:650--662, 2017.

\bibitem{SCI}
Long Ma, Tengyu Ma, Risheng Liu, Xin Fan, and Zhongxuan Luo.
\newblock Toward fast, flexible, and robust low-light image enhancement.
\newblock In {\em Proceedings of the {IEEE/CVF} Conference on Computer Vision
  and Pattern Recognition, New Orleans, LA, USA}, pages 5627--5636, 2022.

\bibitem{Segmentation-Survey}
Shervin Minaee, Yuri Boykov, Fatih Porikli, Antonio Plaza, Nasser Kehtarnavaz,
  and Demetri Terzopoulos.
\newblock Image segmentation using deep learning: {A} survey.
\newblock {\em {IEEE} Trans. Pattern Anal. Mach. Intell.}, 44(7), 2022.

\bibitem{Metric-NIQE}
Anish Mittal, Rajiv Soundararajan, and Alan~C. Bovik.
\newblock Making a "completely blind" image quality analyzer.
\newblock {\em {IEEE} Signal Process. Lett.}, 20(3):209--212, 2013.

\bibitem{DeepDeblur}
Seungjun Nah, Tae~Hyun Kim, and Kyoung~Mu Lee.
\newblock Deep multi-scale convolutional neural network for dynamic scene
  deblurring.
\newblock In {\em Proceedings of the {IEEE} Conference on Computer Vision and
  Pattern Recognition, Honolulu, HI, USA}, pages 257--265, 2017.

\bibitem{LR3M}
Xutong Ren, Wenhan Yang, Wen{-}Huang Cheng, and Jiaying Liu.
\newblock {LR3M:} robust low-light enhancement via low-rank regularized retinex
  model.
\newblock {\em {IEEE} Trans. Image Process.}, 29:5862--5876, 2020.

\bibitem{Deep_Image_Prior}
Dmitry Ulyanov, Andrea Vedaldi, and Victor~S. Lempitsky.
\newblock Deep image prior.
\newblock In {\em Proceedings of the {IEEE} Conference on Computer Vision and
  Pattern Recognition, Salt Lake City, UT, USA}, pages 9446--9454, 2018.

\bibitem{Low_Light_Deraining}
Yecong Wan, Yuanshuo Cheng, Mingwen Shao, and Jordi Gonz{\`{a}}lez.
\newblock Image rain removal and illumination enhancement done in one go.
\newblock {\em Knowl. Based Syst.}, 252:109244, 2022.

\bibitem{Low_Light_Face_Detection_1}
Wenjing Wang, Xinhao Wang, Wenhan Yang, and Jiaying Liu.
\newblock Unsupervised face detection in the dark.
\newblock {\em {IEEE} Trans. Pattern Anal. Mach. Intell.}, 2022.

\bibitem{Low_Light_Face_Detection}
Wenjing Wang, Wenhan Yang, and Jiaying Liu.
\newblock Hla-face: Joint high-low adaptation for low light face detection.
\newblock In {\em Proceedings of the {IEEE} Conference on Computer Vision and
  Pattern Recognition, virtual}, pages 16195--16204, 2021.

\bibitem{LLFlow}
Yufei Wang, Renjie Wan, Wenhan Yang, Haoliang Li, Lap{-}Pui Chau, and Alex~C.
  Kot.
\newblock Low-light image enhancement with normalizing flow.
\newblock In {\em Proceedings of the Thirty-Sixth {AAAI} Conference on
  Artificial Intelligence}, pages 2604--2612, 2022.

\bibitem{Metric-SSIM}
Zhou Wang, Alan~C. Bovik, Hamid~R. Sheikh, and Eero~P. Simoncelli.
\newblock Image quality assessment: from error visibility to structural
  similarity.
\newblock {\em {IEEE} Trans. Image Process.}, 13(4):600--612, 2004.

\bibitem{RetinexNet}
Chen Wei, Wenjing Wang, Wenhan Yang, and Jiaying Liu.
\newblock Deep retinex decomposition for low-light enhancement.
\newblock In {\em Proceedings of the British Machine Vision Conference,
  Newcastle, UK}, page 155, 2018.

\bibitem{Low_Light_Denoising}
Kaixuan Wei, Ying Fu, Yinqiang Zheng, and Jiaolong Yang.
\newblock Physics-based noise modeling for extreme low-light photography.
\newblock {\em {IEEE} Trans. Pattern Anal. Mach. Intell.}, 44(11):8520--8537,
  2022.

\bibitem{DerainCycleGAN}
Yanyan Wei, Zhao Zhang, Yang Wang, Mingliang Xu, Yi Yang, Shuicheng Yan, and
  Meng Wang.
\newblock Deraincyclegan: Rain attentive cyclegan for single image deraining
  and rainmaking.
\newblock {\em {IEEE} Trans. Image Process.}, 30, 2021.

\bibitem{URetinex-Net}
Wenhui Wu, Jian Weng, Pingping Zhang, Xu Wang, Wenhan Yang, and Jianmin Jiang.
\newblock Uretinex-net: Retinex-based deep unfolding network for low-light
  image enhancement.
\newblock In {\em Proceedings of the {IEEE/CVF} Conference on Computer Vision
  and Pattern Recognition, New Orleans, LA, USA}, pages 5891--5900, 2022.

\bibitem{Novel-Captioner}
Yu Wu, Lu Jiang, and Yi Yang.
\newblock Switchable novel object captioner.
\newblock {\em {IEEE} Trans. Pattern Anal. Mach. Intell.}, 45(1), 2023.

\bibitem{FIDE}
Ke Xu, Xin Yang, Baocai Yin, and Rynson W.~H. Lau.
\newblock Learning to restore low-light images via
  decomposition-and-enhancement.
\newblock In {\em Proceedings of the {IEEE/CVF} Conference on Computer Vision
  and Pattern Recognition, Seattle, WA, USA}, pages 2278--2287, 2020.

\bibitem{Rain100H}
Wenhan Yang, Robby~T Tan, Jiashi Feng, Jiaying Liu, Zongming Guo, and Shuicheng
  Yan.
\newblock Deep joint rain detection and removal from a single image.
\newblock In {\em Proceedings of the IEEE conference on computer vision and
  pattern recognition}, pages 1357--1366, 2017.

\bibitem{MPRNet}
Syed~Waqas Zamir, Aditya Arora, Salman~H. Khan, Munawar Hayat, Fahad~Shahbaz
  Khan, Ming{-}Hsuan Yang, and Ling Shao.
\newblock Multi-stage progressive image restoration.
\newblock In {\em Proceedings of the {IEEE} Conference on Computer Vision and
  Pattern Recognition}, pages 14821--14831, 2021.

\bibitem{Weakly--Supervised-Object-Detection}
Dingwen Zhang, Wenyuan Zeng, Jieru Yao, and Junwei Han.
\newblock Weakly supervised object detection using proposal- and semantic-level
  relationships.
\newblock {\em {IEEE} Trans. Pattern Anal. Mach. Intell.}, 44(6):3349--3363,
  2022.

\bibitem{DUAL}
Qing Zhang, Yongwei Nie, and Wei{-}Shi Zheng.
\newblock Dual illumination estimation for robust exposure correction.
\newblock {\em Comput. Graph. Forum}, 38(7):243--252, 2019.

\bibitem{KinD++}
Yonghua Zhang, Xiaojie Guo, Jiayi Ma, Wei Liu, and Jiawan Zhang.
\newblock Beyond brightening low-light images.
\newblock {\em Int. J. Comput. Vis.}, 129(4):1013--1037, 2021.

\bibitem{KinD}
Yonghua Zhang, Jiawan Zhang, and Xiaojie Guo.
\newblock Kindling the darkness: {A} practical low-light image enhancer.
\newblock In {\em Proceedings of the {ACM} International Conference on
  Multimedia, Nice, France}, pages 1632--1640, 2019.

\bibitem{DCC-Net}
Zhao Zhang, Huan Zheng, Richang Hong, Mingliang Xu, Shuicheng Yan, and Meng
  Wang.
\newblock Deep color consistent network for low-light image enhancement.
\newblock In {\em Proceedings of the {IEEE/CVF} Conference on Computer Vision
  and Pattern Recognition, New Orleans, LA, USA}, pages 1889--1898, 2022.

\bibitem{FCL-GAN}
Suiyi Zhao, Zhao Zhang, Richang Hong, Mingliang Xu, Yi Yang, and Meng Wang.
\newblock {FCL-GAN:} {A} lightweight and real-time baseline for unsupervised
  blind image deblurring.
\newblock In {\em Proceedings of the {ACM} International Conference on
  Multimedia, Lisboa, Portugal}, pages 6220--6229, 2022.

\bibitem{CRNet}
Suiyi Zhao, Zhao Zhang, Richang Hong, Mingliang Xu, Haijun Zhang, Meng Wang,
  and Shuicheng Yan.
\newblock Crnet: Unsupervised color retention network for blind motion
  deblurring.
\newblock In {\em Proceedings of the {ACM} International Conference on
  Multimedia, Lisboa, Portugal}, pages 6193--6201, 2022.

\bibitem{Low_Light_Denoising_and_Deblurring}
Yuzhi Zhao, Yongzhe Xu, Qiong Yan, Dingdong Yang, Xuehui Wang, and Lai{-}Man
  Po.
\newblock D2hnet: Joint denoising and deblurring with hierarchical network for
  robust night image restoration.
\newblock In Shai Avidan, Gabriel~J. Brostow, Moustapha Ciss{\'{e}},
  Giovanni~Maria Farinella, and Tal Hassner, editors, {\em Proceedings of the
  17th European Conference, Tel Aviv, Israel}, volume 13667, pages 91--110,
  2022.

\bibitem{Bayes-Unpaired-Restoration}
Dihan Zheng, Xiaowen Zhang, Kaisheng Ma, and Chenglong Bao.
\newblock Learn from unpaired data for image restoration: {A} variational bayes
  approach.
\newblock {\em {IEEE} Trans. Pattern Anal. Mach. Intell.}, 45(5):5889--5903,
  2023.

\bibitem{CycleGAN}
Jun{-}Yan Zhu, Taesung Park, Phillip Isola, and Alexei~A. Efros.
\newblock Unpaired image-to-image translation using cycle-consistent
  adversarial networks.
\newblock In {\em Proceedings of the {IEEE} International Conference on
  Computer Vision, Venice, Italy}, pages 2242--2251, 2017.

\end{thebibliography}
}

\end{document}